\documentclass{ecai} 

\usepackage{latexsym}
\usepackage{amssymb}
\usepackage{amsmath}
\usepackage{amsthm}
\usepackage{booktabs}
\usepackage{enumitem}
\usepackage{graphicx}
\usepackage{color}
\usepackage{booktabs}
\usepackage{multirow}
\usepackage[table,xcdraw]{xcolor}
\usepackage{algorithm2e}
\RestyleAlgo{ruled}
\usepackage{algpseudocode}
\usepackage{caption}
\usepackage{subcaption}
\usepackage{tikz}
\usepackage{hyperref}
\usetikzlibrary{calc} 
\usepackage[most]{tcolorbox}
\usepackage{xcolor}
\usepackage{inconsolata} 

\definecolor{mywhite}{RGB}{255,255,255}
\definecolor{myyellow}{RGB}{255,255,200}
\definecolor{myorange}{RGB}{255,220,180}

\newtcolorbox{examplebox}[1][]{
  enhanced,
  top=1mm,
  bottom=1mm,
  left=1mm,
  right=1mm,
  colback=mywhite,
  colframe=myorange,
  coltitle=black,
  boxrule=1pt,
  arc=4pt,
  colbacklower=myyellow,
  title=,
  fonttitle=\bfseries,
  before upper={\ttfamily\scriptsize\vspace{1pt}}, 
  after upper={\vspace{1pt}}
}

\newcommand{\BibTeX}{B\kern-.05em{\sc i\kern-.025em b}\kern-.08em\TeX}
\newcommand{\mtok}{\langle m \rangle}

\begin{document}


\begin{frontmatter}


\paperid{6946} 


\title{Improving Text Style Transfer using Masked Diffusion Language Models with Inference-time Scaling}


\author[A]{\fnms{Tejomay Kishor}~\snm{Padole}\thanks{Corresponding Author. Email: tejomaypadole@cse.iitb.ac.in}
}
\author[A]{\fnms{Suyash P.}~\snm{Awate}}
\author[A]{\fnms{Pushpak }~\snm{Bhattacharyya}} 

\address[A]{Dept. of Computer Science and Engineering, Indian Institute of Technology, Bombay}


\begin{abstract}
Masked diffusion language models (MDMs) have recently gained traction as a viable generative framework for natural language. This can be attributed to its scalability and ease of training compared to other diffusion model paradigms for discrete data, establishing itself as the state-of-the-art non-autoregressive generator for discrete data. Diffusion models, in general, have shown excellent ability to improve the generation quality by leveraging inference-time scaling either by increasing the number of denoising steps or by using external verifiers on top of the outputs of each step to guide the generation. In this work, we propose a verifier-based inference-time scaling method that aids in finding a better candidate generation during the denoising process of the MDM. Our experiments demonstrate the application of MDMs for standard text-style transfer tasks and establish MDMs as a better alternative to autoregressive language models. Additionally, we show that a simple soft-value-based verifier setup for MDMs using off-the-shelf pre-trained embedding models leads to significant gains in generation quality even when used on top of typical classifier-free guidance setups in the existing literature. 
\end{abstract}

\end{frontmatter}


\begin{figure*}[t]
\centering
\vspace{0.1cm}
\includegraphics[width=18cm]{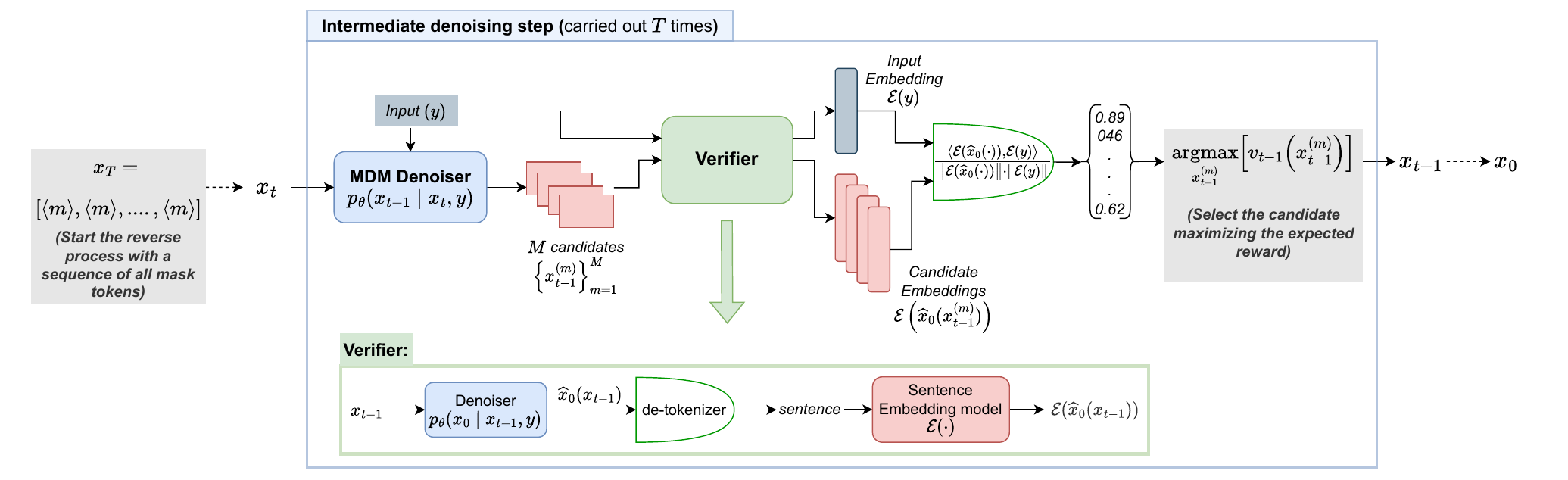}
\caption{Diagram showing a single denoising step (i.e., the reverse process) of MDM with the soft-value diffusion decoding method paired with a pre-trained sentence-embedding-based verifier at an arbitrary timestep $ t$. An intermediate noisy sample $ x_{t}$ is passed through the denoiser, and $M$ candidates for $ x_{t-1} $ are sampled. The $M$ candidates pass through the verifier which follows three steps: \textbf{(1)} $ x_0-\text{prediction}$ for each candidate, \textbf{(2)} detokenize $ x_0$ to form a sentence, and \textbf{(3)} pass the sentence through the embedding model to obtain sentence embedding. The input sentence embedding and the candidate embeddings are used to calculate the value function $ v_{t-1}(\cdot) $ (i.e., the expected reward as defined in \ref{subsec: svdd}). The final $ x_{t-1} $ is chosen such that it maximizes the value function.} 
\vspace{0.2cm}
\label{fig: diagram}
\end{figure*}

\section{Introduction}
The current landscape of natural language generation is dominated by pre-trained autoregressive large language models (LLMs). The next-token prediction strategy these models employ has proven to be very effective for generative modeling on language data that is typically provided with large-scale corpora \citep{gpt3, llama3}. However, sequential sampling also comes with limitations. Specifically, such a strategy suffers from "sampling drifts", meaning that the generation degrades as it progresses due to error accumulation \cite{sedd}. Diffusion language modeling offers an alternative strategy by proposing non-autoregressive sampling of tokens with iterative denoising, potentially mitigating these downsides. Recently, the masked diffusion language modeling (MDM) paradigm has been proposed as a scalable approach that has been quite competitive with autoregressive models on the same parameter scale \citep{mdlm, smdm, llada}. 

Apart from non-autoregressive generation, diffusion models have also demonstrated excellent abilities in guiding the generation towards specific outputs. Diffusion guidance techniques can be mainly classified as: \textbf{(1) classifier guidance} \cite{classifier_guidance} where an external classifier is trained, which provides gradient-based guidance signals to the diffusion model during sampling, \textbf{(2) classifier-free guidance (CFG)} \cite{cfg} where a diffusion model is trained jointly for conditional and unconditional generation which improves fidelity during sampling, and \textbf{(3) training-free guidance (TFG)} \cite{tfg} where off-the-shelf pre-trained models are used directly to provide gradient-based guidance signals to a diffusion model. Apart from these, \textbf{derivative-free guidance} techniques have recently emerged as an alternative to classifier guidance and TFG, eliminating the gradient-computation step during sampling and easing the adaptation of off-the-shelf pre-trained models for guidance. Such techniques either follow a sequential Monte Carlo approach to improve the conditioning \cite{dps-smc} or perform explicit reward maximization by setting the pre-trained model's likelihood/ classifier logits as rewards \cite{svdd}.

Guidance mechanisms in diffusion models are effective methods to scale the inference-time compute in diffusion models \cite{compute_scaling_diffusion}. For instance, external models (either trained specifically for a task or used off-the-shelf) that offer guidance signals to diffusion models are seen as verifiers that consume extra compute. In return, they provide more control and potentially improve generation quality when combined with search algorithms. Existing work on diffusion models for discrete data has proposed methods to apply guidance mechanisms in discrete spaces \citep{simple_discrete_guidance, unlocking_discrete_guidance, svdd}. However, their experiments are mostly restricted to discrete data modalities such as small molecules, DNA sequences, and protein sequences, while the language modality remains largely unexplored. The work of \cite{debiasing_guidance} showcases discrete diffusion guidance on language with sequential Monte Carlo but only applies it to attribute control for unconditional text generation. The work of \cite{diff_style_transfer} is conceptually closest to ours, which explores style transfer with seq-to-seq continuous diffusion \cite{diffuseq}, but they do not explore any guidance mechanisms or discuss inference-time scaling with external verifier-based search algorithms.

In this paper, we propose verifier-based inference-time scaling with masked diffusion language models (MDMs) for text-style transfer, one of the well-studied and important conditional language generation tasks. Specifically, we design a novel derivative-free guidance method with MDMs using pre-trained sentence-embedding models as external verifiers. We demonstrate the effectiveness of this setup using two standard text-style transfer tasks. To the best of our knowledge, we are the first to propose this setting for any conditional language generation task.\footnote{Code can be found \href{https://github.com/Tejomay128/MDM_style_transfer}{here}.} Our contributions are
\vspace{-2pt}
\begin{enumerate}
    \item a novel derivative-free guidance method for MDMs using guidance signals from off-the-shelf pre-trained sentence embedding models to enhance conditional language generation. (Section \ref{subsec: svdd}).

    \item demonstration (for the first time, to the best of our knowledge) of the utility of MDMs for text style transfer (with available parallel data for training) by performing seq-to-seq fine-tuning on pre-trained MDMs (Section \ref{sec: methodology}).

    \item extensive experiments with autoregressive and diffusion baselines establishing MDMs as a better alternative and highlighting the impact of inference-time scaling in MDMs (Section \ref{sec: results}). 
    
    \item empirical analysis of inference-time compute scaling methods along different scaling axes, namely denoising steps and number of samples considered during the verification step to obtain guidance signals (Section \ref{sec: test-time compute}).
\end{enumerate}

\section{Background}

\subsection{Text Style Transfer}
Text style transfer (TST) is a natural language generation task that involves modifying the stylistic properties of a given input text (e.g., sentiment, formality, or dialect) while preserving its underlying semantic content. TST is considered an important task in NLP since the choice of style plays a vital role in understanding user intent \cite{tst-survey}. Formally, given a source sentence \( x \) with style \( s_1 \), the goal of TST is to generate a new sentence \( x' \) such that \( x' \approx x \) in content and exhibits a target style \( s_2 \). The task has been explored in various supervised, unsupervised, and semi-supervised settings. When parallel corpora are available---that is, sentence pairs \( (x, y) \) where \( x \) is a sentence in style \( s_1 \) and \( y \) is its corresponding rewriting in style \( s_2 \) with the same semantic content---the task can be cast as a supervised sequence-to-sequence learning problem.

\subsection{Masked Diffusion Language Models} \label{subsec: MDM}
Masked Diffusion Language Models (MDMs) define the diffusion forward and reverse processes in the discrete (or the token) space \citep{multinomial_diffusion, austin_diffusion}. The forward process iteratively replaces tokens in the sequence with mask tokens and the reverse process is learned to iteratively unmask tokens to generate new samples \citep{mdlm, smdm}. 

Let $x_0 = [x_0^{0}, x_0^{1}, ...., x_0^{L-1}]$ be a sentence where $L$ is the length of the sentence. Let $t \in [0,1]$ be the timestep for the diffusion process. Let $V$ be the total vocabulary size. The forward process is defined as:

\begin{align}
    q_{t|0}(x_t | x_0) &= \prod_{i=0}^{L-1} q_{t|0}(x_t^i | x_0^i) \;\; \text{ and} \notag \\
    q_{t|0}(x_t^i | x_0^i) &:= 
    \begin{cases}
        \alpha_t, &x_t^i = x_0^i, \\
        1 - \alpha_t, &x_t^i = \mtok,
    \end{cases} \label{eqn: forward_process}
\end{align}

where $x_t$ is the sequence at timestep $t$, $\alpha_t$ is the hyperparameter that controls the noise level (i.e., the proportion of masked tokens in the sequence) at a given timestep, $\mtok$ represents the mask token and $ q_0(\cdot) $ is the data distribution. Note that the forward process is factorized across the sequence length, meaning that the noise is added independently to each token in the sequence. For timesteps $s$ and $t$ with $ 0 \leq s < t \leq 1 $, the corresponding reverse process is defined as:

\begin{align}
    q_{s|t}(x_s | x_t) &= \prod_{i=0}^{L-1} q_{s|t}(x_s^i | x_t) \;\; \text{ and} \notag \\
    q_{s|t}(x_s^i | x_t) &=
    \begin{cases}
        \frac{t-s}{t} q_{0|t}(x_s^i | x_t), &x_t^i = \mtok, x_s^i \neq \mtok \\
        \frac{s}{t}, &x_t^i = \mtok, x_s^i = \mtok \\
        1, &x_t^i \neq \mtok, x_s^i = x_t^i \\
    \end{cases} \label{eqn: reverse_process}
\end{align}

The distribution $q_{0|t}(\cdot| \cdot)$ is learned by parameterizing it as an output of a neural network. Intuitively, a neural network is learned to predict all the masked tokens given the unmasked tokens in a noisy sentence at all possible noise levels. Formally, we write the estimated data distribution as $p_{\theta}(x_0)$ and the neural network estimate as $ p_{\theta}(\cdot|x_t) \approx q_{0|t}(\cdot | x_t)$ where $\theta$ are the parameters of the neural network. To learn the network, \citep{shi2024simplified, mdlm} derive a simplified upper-bound on the negative log-likelihood as the training objective:

\begin{align}
  -\log p_{\theta}(x_0) \leq \int_{0}^1 \frac{\alpha'_t}{1-\alpha_t} \;\mathbb{E} \left [ \underset{x_t^i = \mtok}{\sum_{i}} - \log p_{\theta}( x_0^i|x_t ) \right] dt\label{eqn: loss}
\end{align}

The expectation is under the forward distribution $ q(x_t | x_0) $. The objective is similar to the masked language modeling objective \citep{bert} since it computes the cross-entropy loss on the masked positions in the noisy sentence. In practice, MDMs have proven their ability to scale easily \cite{smdm, llada} to larger sizes, decreasing the performance gap between text diffusion models and autoregressive LMs. Additionally, unlike typical diffusion models, masked diffusion processes can be learned with a time-independent neural network, which allows us to use standard Transformer architectures with minimal changes to parameterize the reverse process.

\section{Methodology} \label{sec: methodology}
Our base method follows from section \ref{subsec: MDM}. We take an MDM pre-trained on language data and fine-tune the model on the downstream style transfer tasks. Our fine-tuning method is tailored for sequence-to-sequence learning setups. Let $ x_0 $ be the target sequence of length $L_1$ (following notations from section \ref{subsec: MDM}) and $y$ be the input conditional sequence of length $L_2$. During training, we concatenate $ x_0 $ and $y$ to form a training instance $ x^{\text{inp}} = [y^1, y^2, ...., y^{L_2},\text{<sep>}, x_0^1, x_0^2, ...., x_0^{L_1}] $, where <sep> is either a separator token or a string. 

To perform the forward diffusion process, we add noise to the target part of $ x^{\text{inp}} $ (i.e., $x_0$) by following equation \ref{eqn: forward_process} to form $ x_t^{\text{inp}} $ by randomly sampling a timestep $t$ from a uniform random distribution $ \mathcal{U}(0, 1) $. We keep the conditional part of $ x^{\text{inp}} $ (i.e., $y$) intact during the forward process \citep{diffuseq, smdm}. We set $ \alpha_t = 1-t $ following prior work \citep{sedd, shi2024simplified, mdlm}.  Next, we pass $ x_t^{\text{inp}} $ through the neural network $ \text{NN}_{\theta} $ (which means we have $ p_{\theta}(x_0| x_t, y) = \text{NN}_{\theta}(x_t^{\text{inp}}) $). The denoiser $ \text{NN}_{\theta} $ is a transformer with bi-directional self-attention. Following equation \ref{eqn: loss}, we compute the loss values on the masked positions. We always compute loss only on the target sequence tokens since we only add noise to the target sequence, which makes this similar to how autoregressive LMs are instruction-tuned. More details on the training and sampling algorithms for MDMs are described in Appendix A.

\begin{table*}[t]
    \centering
    \resizebox{16cm}{!}{%
    \begin{tabular}{@{}ccccccccl@{}}
    \toprule
    \multicolumn{9}{c}{\textbf{WikiLarge complicated to simple text style transfer}}                                                                                                                                                                                                                                                                                                                                                                                                                                                                                                                                                                                            \\ \midrule
    {\color[HTML]{000000} \textbf{Model}}                                                                           & {\color[HTML]{000000} \textbf{\#params}}      & {\color[HTML]{000000} \textbf{\#steps}} & {\color[HTML]{000000} \textbf{BLEU $\uparrow$}}                       & {\color[HTML]{000000} \textbf{}}                                               & {\color[HTML]{000000} \textbf{SARI $\uparrow$}}                       & {\color[HTML]{000000} \textbf{}}                                               & {\color[HTML]{000000} \textbf{LENS $\uparrow$}}                       & \multicolumn{1}{c}{\textbf{}}                                         \\ \midrule
    {\color[HTML]{000000} \textbf{}}                                                                                & {\color[HTML]{000000} \textbf{}}              & {\color[HTML]{000000} \textbf{}}        & {\color[HTML]{000000} \textbf{w/o SVDD}}                              & {\color[HTML]{000000} \textbf{with SVDD (ours)}}                                      & {\color[HTML]{000000} \textbf{w/o SVDD}}                              & {\color[HTML]{000000} \textbf{with SVDD (ours)}}                                      & {\color[HTML]{000000} \textbf{w/o SVDD}}                              & \multicolumn{1}{c}{\textbf{with SVDD (ours)}}                                \\
    {\color[HTML]{000000} }                                                                                         & {\color[HTML]{000000} }                       & {\color[HTML]{000000} 8}                & {\color[HTML]{000000} $ 54.609_{\pm 0.974} $}                         & \cellcolor[HTML]{FFFFC7}{\color[HTML]{000000} $ 60.923_{\pm 0.617} $}          & {\color[HTML]{000000} $ 33.653_{\pm 0.252} $}                         & \cellcolor[HTML]{FFFFC7}{\color[HTML]{000000} $ 35.500_{\pm 0.249} $}          & {\color[HTML]{000000} $ 18.794_{\pm 0.804} $}                         & \cellcolor[HTML]{FFFFC7}{\color[HTML]{000000} $ 22.118_{\pm 0.733} $} \\
    {\color[HTML]{000000} }                                                                                         & {\color[HTML]{000000} }                       & {\color[HTML]{000000} 16}               & {\color[HTML]{000000} $ 62.168_{\pm 0.608} $}                         & \cellcolor[HTML]{FFFFC7}{\color[HTML]{000000} $ 67.055_{\pm 0.682} $}          & {\color[HTML]{000000} $ 35.870_{\pm 0.157} $}                         & \cellcolor[HTML]{FFFFC7}{\color[HTML]{000000} $ 37.385_{\pm 0.202} $}          & {\color[HTML]{000000} $ 28.552_{\pm 0.765} $}                         & \cellcolor[HTML]{FFFFC7}{\color[HTML]{000000} $ 31.354_{\pm 0.644} $} \\
    \multirow{-3}{*}{{\color[HTML]{000000} \textbf{\begin{tabular}[c]{@{}c@{}}MDM w/o CFG\\ \end{tabular}}}}  & {\color[HTML]{000000} }                       & {\color[HTML]{000000} 64}               & {\color[HTML]{000000} $ 41.467_{\pm 0.629} $}                         & \cellcolor[HTML]{FFFFC7}{\color[HTML]{000000} $ 44.598_{\pm 0.763} $}          & {\color[HTML]{000000} $ 31.591_{\pm 0.187} $}                         & \cellcolor[HTML]{FFFFC7}{\color[HTML]{000000} $ 32.437_{\pm 0.219} $}          & {\color[HTML]{000000} $ 28.983_{\pm 0.552} $}                         & \cellcolor[HTML]{FFFFC7}{\color[HTML]{000000} $ 30.658_{\pm 0.687} $} \\ \cmidrule(l){3-9} 
    {\color[HTML]{000000} }                                                                                         & {\color[HTML]{000000} }                       & {\color[HTML]{000000} 8}                & \cellcolor[HTML]{FFCCC9}{\color[HTML]{000000} $ 69.545_{\pm 0.431} $} & \cellcolor[HTML]{FFCE93}{\color[HTML]{000000} $ 71.251_{\pm 0.327} $}          & \cellcolor[HTML]{FFCCC9}{\color[HTML]{000000} $ 44.089_{\pm 0.205} $} & \cellcolor[HTML]{FFCE93}{\color[HTML]{000000} $ 44.864_{\pm 0.139} $}          & \cellcolor[HTML]{FFCCC9}{\color[HTML]{000000} $ 31.623_{\pm 0.528} $} & \cellcolor[HTML]{FFCE93}{\color[HTML]{000000} $ 33.258_{\pm 0.436} $} \\
    {\color[HTML]{000000} }                                                                                         & {\color[HTML]{000000} }                       & {\color[HTML]{000000} 16}               & \cellcolor[HTML]{FFCCC9}{\color[HTML]{000000} $ 78.854_{\pm 0.633} $} & \cellcolor[HTML]{FFCE93}{\color[HTML]{000000} $ 80.208_{\pm 0.469} $}          & \cellcolor[HTML]{FFCCC9}{\color[HTML]{000000} $ 47.270_{\pm 0.215} $} & \cellcolor[HTML]{FFCE93}{\color[HTML]{000000} $ 47.949_{\pm 0.160} $}          & \cellcolor[HTML]{FFCCC9}{\color[HTML]{000000} $ 40.133_{\pm 0.505} $} & \cellcolor[HTML]{FFCE93}{\color[HTML]{000000} $ 41.650_{\pm 0.381} $} \\
    \multirow{-3}{*}{{\color[HTML]{000000} \textbf{\begin{tabular}[c]{@{}c@{}}MDM with CFG\\ \end{tabular}}}} & \multirow{-6}{*}{{\color[HTML]{000000} 162M}} & {\color[HTML]{000000} 64}               & \cellcolor[HTML]{FFCCC9}{\color[HTML]{000000} $ 86.567_{\pm 0.419} $} & \cellcolor[HTML]{FFCE93}{\color[HTML]{000000} $ \mathbf{87.403_{\pm 0.374}} $} & \cellcolor[HTML]{FFCCC9}{\color[HTML]{000000} $ 48.578_{\pm 0.136} $} & \cellcolor[HTML]{FFCE93}{\color[HTML]{000000} $ \mathbf{49.159_{\pm 0.179}} $} & \cellcolor[HTML]{FFCCC9}{\color[HTML]{000000} $ 46.812_{\pm 0.612} $} & \cellcolor[HTML]{FFCE93}{\color[HTML]{000000} $ 47.999_{\pm 0.534}$}  \\ \cmidrule(l){3-9} 
    \textbf{GENIE}                                                                                                  & 144M                                          & 64                                      & \multicolumn{2}{c}{$ 71.842_{\pm 0.774} $}                                                                                                             & \multicolumn{2}{c}{$ 43.268_{\pm 0.421} $}                                                                                                             & \multicolumn{2}{c}{$ \mathbf{53.863_{\pm 0.868}} $}                                                                                           \\ \cmidrule(l){3-9} 
    {\color[HTML]{000000} }                                                                                         & {\color[HTML]{000000} 136M}                   & {\color[HTML]{000000} \textbf{-}}       & \multicolumn{2}{c}{{\color[HTML]{000000} $ 43.202_{\pm 1.465} $}}                                                                                      & \multicolumn{2}{c}{{\color[HTML]{000000} $ 34.186_{\pm 0.608} $}}                                                                                      & \multicolumn{2}{c}{{\color[HTML]{000000} $ 43.126_{\pm 1.078} $}}                                                                             \\ \cmidrule(l){3-9} 
    \multirow{-2}{*}{{\color[HTML]{000000} \textbf{SmolLM2}}}                                                       & {\color[HTML]{000000} 360M}                   & {\color[HTML]{000000} \textbf{-}}       & \multicolumn{2}{c}{{\color[HTML]{000000} $ 50.148_{\pm 3.136} $}}                                                                                      & \multicolumn{2}{c}{{\color[HTML]{000000} $ 38.748_{\pm 2.161} $}}                                                                                      & \multicolumn{2}{c}{{\color[HTML]{000000} $ 48.348_{\pm 1.874} $}}                                                                             \\ \cmidrule(l){3-9} 
    {\color[HTML]{000000} \textbf{Qwen2.5}}                                                                         & {\color[HTML]{000000} 490M}                   & {\color[HTML]{000000} \textbf{-}}       & \multicolumn{2}{c}{{\color[HTML]{000000} $ 52.268_{\pm 1.573} $}}                                                                                      & \multicolumn{2}{c}{{\color[HTML]{000000} $ 40.463_{\pm 0.722} $}}                                                                                      & \multicolumn{2}{c}{{\color[HTML]{000000} $ 49.865_{\pm 1.053} $}}                                                                             \\ \bottomrule
    \end{tabular}%
    }   
    \vspace{0.2cm}
    \caption{Main results for the Wikilarge simplicity style transfer dataset. As mentioned in section \ref{subsec: metrics}, we report the mean and the standard deviation (in the subscript) across 20 independent sampling runs. $\uparrow$ indicates "higher the better" for the metric. The non-coloured cells report baseline scores, while the coloured cells represent different inference-time scaling settings.}
    \vspace{0.3cm}
    \label{table: wikilarge}
\end{table*}

\subsection{Classifier-Free Guidance}
Our training methodology involves using classifier-free guidance (CFG) \cite{cfg}, which proposes to train the neural network to estimate both conditional (i.e., $ p_{\theta}(x_0 | x_t, y) $) and unconditional (i.e., $ p_{\theta}(x_0 | x_t) $) distributions of the clean data. For learning on text data, this can be achieved by randomly setting the conditional sequence $y$ to $ \phi = [\mtok, \mtok, ...., \mtok]$ (i.e., a sequence of all mask tokens of the same length as $y$) during training \cite{simple_discrete_guidance}. This setting is equivalent to modeling the unconditional distribution, i.e., $ p_{\theta}(x_0 | x_t) = p_{\theta}(x_0 | x_t, \phi) $. 

After training, these distributions are used during inference to estimate the conditional log-probability of the clean data $x_0$ as follows:

\begin{align}
    \log p_{\theta, \gamma}(x_0 | x_t, y) = \; &\gamma \,\log p_{\theta}(x_0 | x_t, y) \; \\ \notag &+ \\ \notag 
    &(1-\gamma) \, \log p_{\theta}(x_0 | x_t) + c. \label{eqn: cfg}
\end{align}

Here, $ \gamma $ is the classifier-free guidance scale, which offers a trade-off between sample diversity and fidelity. The constant $c$ can be ignored since we compute probabilities using the softmax function, the results of which remain unaffected by the value of $c$. Typically, higher values of $ \gamma $ (> 1) provide more grounded outputs and result in better sample quality. At $\gamma = 1$, the estimation boils down to the conditional estimate without any CFG.

\subsection{Soft-Value Diffusion Decoding} \label{subsec: svdd}
The generative process of the diffusion models can be scaled beyond denoising steps by designing an evaluation mechanism for the outputs at each step. Using these external verification signals provides guidance to the generation process towards potentially better outputs. \cite{compute_scaling_diffusion} denotes this as scaling along the verifier axis. In this work, we propose to design the verification step for TST tasks using off-the-shelf pretrained sentence embedding models. Specifically, we score (or reward) a sampled candidate from the denoiser at each denoising step using the cosine similarity between the candidate embeddings and the input sentence embeddings. The idea is to reward the capturing of the semantic content of the input sentence in the style-transferred candidate generated from the model. Formally, the reward is defined as
\begin{align}
    R_{\mathcal{E}, t} \left(\widehat{x}_0 (x_t), y \right) := \frac{\langle \mathcal{E}(\widehat{x}_0(x_t)), \mathcal{E}(y) \rangle}{\left \|\mathcal{E}(\widehat{x}_0(x_t))  \right \| \cdot \left \|\mathcal{E}(y)  \right \|}, 
\end{align}
where $ \mathcal{E(\cdot)} $ is an off-the-shelf pre-trained sentence embedding model and $ \widehat{x}_0(x_t) \sim p_{\theta}(x_0 | x_t, y) $. Along with adding an external verifier for evaluation, a search algorithm needs to be in place to find the best candidate at every denoising step based on the output from the verifier. Using the scores defined above as a reward function, we implement the \textbf{Soft-Value Diffusion Decoding (SVDD)} algorithm \cite{svdd}. Figure \ref{fig: diagram} explains how the algorithm works with a single intermediate step of the reverse diffusion process. The algorithm derives from sampling while maximizing the reward, which boils down to sampling multiple possible candidates during a denoising step and choosing the best candidate that results in the maximum expected reward at that step. With a large enough pool of candidates, this results in an explicit reward maximization at the end of the denoising trajectory. 
Let $ v_{t}(\cdot) $ be the value function denoting the expected reward at $ t=0 $ from an arbitrary noisy state at step $ t' $. Following the \textbf{posterior-mean-approximation (PMA)} method from \cite{svdd}, the value function induced by $ x_{t'} $ can be approximated as $ v_{t'}(x_{t'}) \approx R_{\mathcal{E},t'}(\widehat{x}_0(x_{t'}), y) $. To estimate the optimal denoising route, at every timestep, we draw $M$ (> 1) independent $ x_{t'} $samples and select the $ x_{t'} $ that maximizes the value function. Algorithm \ref{alg: svdd} details the SVDD method explained above. Refer to Appendix \ref{appendix B} for more details on how SVDD with PMA leads to reward maximization.

\begin{algorithm}[t]
    \caption{The MDM reverse process with SVDD} \label{alg: svdd}
    
    \textbf{Given:} Denoiser $ \text{NN}_{\theta} $, sentence embedding model $\mathcal{E}(\cdot)$, input/source sentence $y$, the value function $ v_t(\cdot) = R_{\mathcal{E}, t} \left(\cdot, y \right) $ and total denoising steps $ T $.  \\
    \textbf{Initialize:} $ t = T$ \\
    \While{$ t \geq 1  $}{
        Compute $ p_{\theta}(x_0 | x_t, y) = \text{NN}_{\theta}(x_t^{\text{inp}}) \; (\approx q_{0|t}(x_0|x_t, y)) $ \\
        Compute $ p_{\theta}(x_{t-1}|x_t, y) $ following eqn. \ref{eqn: reverse_process} \\
        Sample $M$ candidates: $ \{x_{t-1}^{(m)}\}_{m=1}^M \sim p_{\theta}(x_{t-1}|x_t, y)$ \\
        $ x_{t-1} = \underset{x_{t-1}^{(m)}}{\text{argmax}} \left[ v_{t-1}(x_{t-1}^{(m)}) \right], \text{where } m \in \{1,2,...,M\}$ \\
        $ t = t-1 $
    }
    \textbf{return} $x_0$
\end{algorithm}

The above formulation provides two advantages:
\begin{enumerate}
    \item As long as the denoiser is available, there is no need to explicitly train a separate reward model (i.e., the verifier, which is a pre-trained sentence embedding model in our case). This is because we have an estimate of the clean data $ x_0 $ at every denoising step, which can be directly passed to the reward model for evaluation of the expected reward.

    \item Since the method is derivative-free, it gives the flexibility of choosing any pre-trained sentence embedding model. This is not possible with derivative-based guidance methods, where the choice of the sentence embedding model would be restricted by the requirement of having the same tokenizer as the denoiser.
\end{enumerate}

\section{Experimental Details}

\subsection{Datasets}
We train and evaluate the models on two TST datasets: \textbf{(1) WikiLarge simplicity style transfer dataset}\footnote{\url{https://github.com/XingxingZhang/dress?tab=readme-ov-file}} \cite{wikilarge} and \textbf{(2) Bible prose style transfer dataset}\footnote{\url{https://github.com/keithecarlson/StyleTransferBibleData}} \cite{bible}. The simplicity style transfer dataset consists of sentences and their simplified version. The corresponding training/validation/test split in the dataset is 296K/1K/0.35K\footnote{The test set consists of 8 references per input which can be accessed from \url{https://github.com/cocoxu/simplification/tree/master/data/turkcorpus}}, respectively. The training data of the Bible prose style transfer consists of 1.5M sentence pairs representing conversions between different versions of biblical sentences as well as bible sentences in basic English style. The corresponding validation set consists of around 91K sentence pairs. For evaluation, we consider two standard test sets: conversion of public version biblical sentences to basic English sentences (PUB-BBE) and conversion of public version biblical sentences to the American Standard Version (PUB-ASV). Each of the test sets consists of around 12K sentence pairs.   

\subsection{Baseline Methods}
We compare our method with the following baselines:
\begin{enumerate}
    \item \textbf{Autoregressive Baselines}:
    \begin{itemize}
        \item \textbf{SmolLM2 \cite{smollm2}:} a family of lightweight autoregressive LMs pre-trained on 2T tokens. We perform seq-to-seq fine-tuning on the 135M and 360M parameter models from the family to create the baseline.

        \item \textbf{Qwen 2.5 \cite{qwen2.5}:} a family of autoregressive LMs pre-trained on 18T tokens.  We perform seq-to-seq fine-tuning on the 490M parameter model from the family to create the baseline.
    \end{itemize}
    
    \item \textbf{GENIE \cite{GENIE}:} a continuous-space diffusion model that is pre-trained on 160Gb of news, books, stories, and web text. Unlike MDMs, GENIE performs the diffusion processes in the embeddings of the tokens instead of directly on the discrete token sequence.

    \item \textbf{MDM without inference-time scaling:} generate outputs from the fine-tuned MDM without inference-time scaling, i.e., without using classifier-free guidance or SVDD or both.
\end{enumerate}

In addition to these baselines, we compare multiple different settings during inference with the fine-tuned MDM, the details of which are discussed in section \ref{subsec: ablations}.

\begin{table*}[t]
\centering
    \resizebox{17cm}{!}{%
    \begin{tabular}{@{}ccccccclccc@{}}
    \toprule
    \multicolumn{11}{c}{\cellcolor[HTML]{FFFFFF}\textbf{Public Versions of Bible to Bible in Basic English (PUB-BBE)}}                                                                                                                                                                                                                                                                                                                                                                                                                                                               \\ \midrule
    \textbf{Model}                                                                           & \textbf{\#params}      & \textbf{\#steps} & \multicolumn{2}{c}{\textbf{BLEU $\uparrow$}}                                                             & \multicolumn{2}{c}{\textbf{ROUGE-L $\uparrow$}}                                                          & \multicolumn{2}{c}{\textbf{METEOR $\uparrow$}}                                                           & \multicolumn{2}{c}{\textbf{BERTScore $\uparrow$}}                                                        \\ \midrule
                                                                                             &                        &                  & \textbf{w/o SVDD}                              & \textbf{with SVDD (ours)}                                      & \textbf{w/o SVDD}                              & \textbf{with SVDD (ours)}                                      & \multicolumn{1}{c}{\textbf{w/o SVDD}}          & \textbf{with SVDD (ours)}                                      & \textbf{w/o SVDD}                              & \multicolumn{1}{l}{\textbf{with SVDD (ours)}}                  \\
                                                                                             &                        & 8                & $ 16.030_{\pm 0.046} $                         & \cellcolor[HTML]{FFFFC7}$ 17.440_{\pm 0.040} $          & $ 44.113_{\pm 0.051} $                         & \cellcolor[HTML]{FFFFC7}$ 45.642_{\pm 0.039} $          & $ 39.073_{\pm 0.069} $                         & \cellcolor[HTML]{FFFFC7}$ 40.937_{\pm 0.031} $          & $ 87.502_{\pm 0.025} $                         & \cellcolor[HTML]{FFFFC7}$ 87.980_{\pm 0.015} $          \\
                                                                                             &                        & 16               & $ 22.687_{\pm 0.036} $                         & \cellcolor[HTML]{FFFFC7}$ 24.435_{\pm 0.032} $          & $ 50.630_{\pm 0.038} $                         & \cellcolor[HTML]{FFFFC7}$ 52.276_{\pm 0.063} $          & $ 46.574_{\pm 0.049} $                         & \cellcolor[HTML]{FFFFC7}$ 48.699_{\pm 0.054} $          & $ 89.715_{\pm 0.025} $                         & \cellcolor[HTML]{FFFFC7}$ 90.205_{\pm 0.035} $          \\
    \multirow{-3}{*}{\textbf{\begin{tabular}[c]{@{}c@{}}MDM w/o CFG\\ \end{tabular}}}  &                        & 64               & $ 26.343_{\pm 0.047} $                         & \cellcolor[HTML]{FFFFC7}$ 27.494_{\pm 0.052} $          & $ 52.937_{\pm 0.042} $                         & \cellcolor[HTML]{FFFFC7}$ 54.176_{\pm 0.053} $          & $ 48.955_{\pm 0.055} $                         & \cellcolor[HTML]{FFFFC7}$ 50.439_{\pm 0.063} $          & $ 90.715_{\pm 0.030} $                         & \cellcolor[HTML]{FFFFC7}$ 91.118_{\pm 0.021} $          \\ \cmidrule(l){3-11} 
                                                                                             &                        & 8                & \cellcolor[HTML]{FFCCC9}$ 18.637_{\pm 0.030} $ & \cellcolor[HTML]{FFCE93}$ 19.923_{\pm 0.056} $          & \cellcolor[HTML]{FFCCC9}$ 45.489_{\pm 0.034} $ & \cellcolor[HTML]{FFCE93}$ 46.901_{\pm 0.056} $          & \cellcolor[HTML]{FFCCC9}$ 41.839_{\pm 0.051} $ & \cellcolor[HTML]{FFCE93}$ 43.648_{\pm 0.077} $          & \cellcolor[HTML]{FFCCC9}$ 88.331_{\pm 0.020} $ & \cellcolor[HTML]{FFCE93}$ 88.751_{\pm 0.024} $          \\
                                                                                             &                        & 16               & \cellcolor[HTML]{FFCCC9}$ 25.870_{\pm 0.044} $ & \cellcolor[HTML]{FFCE93}$ 27.058_{\pm 0.040} $          & \cellcolor[HTML]{FFCCC9}$ 52.340_{\pm 0.043} $ & \cellcolor[HTML]{FFCE93}$ 53.480_{\pm 0.033} $          & \cellcolor[HTML]{FFCCC9}$ 50.555_{\pm 0.047} $ & \cellcolor[HTML]{FFCE93}$ 52.059_{\pm 0.044} $          & \cellcolor[HTML]{FFCCC9}$ 90.476_{\pm 0.012} $ & \cellcolor[HTML]{FFCE93}$ 90.804_{\pm 0.026} $          \\
    \multirow{-3}{*}{\textbf{\begin{tabular}[c]{@{}c@{}}MDM with CFG\\ \end{tabular}}} & \multirow{-6}{*}{162M} & 64               & \cellcolor[HTML]{FFCCC9}$ 32.704_{\pm 0.031} $ & \cellcolor[HTML]{FFCE93}$ \mathbf{33.304_{\pm 0.029}} $ & \cellcolor[HTML]{FFCCC9}$ 58.187_{\pm 0.038} $ & \cellcolor[HTML]{FFCE93}$ \mathbf{58.738_{\pm 0.023}} $ & \cellcolor[HTML]{FFCCC9}$ 57.572_{\pm 0.058} $ & \cellcolor[HTML]{FFCE93}$ \mathbf{58.305_{\pm 0.040}} $ & \cellcolor[HTML]{FFCCC9}$ 92.371_{\pm 0.024} $ & \cellcolor[HTML]{FFCE93}$ \mathbf{92.532_{\pm 0.017}} $ \\ \cmidrule(l){3-11} 
    \textbf{GENIE}                                                                           & 144M                   & 64               & \multicolumn{2}{c}{$ 19.275_{\pm 0.064} $}                                                               & \multicolumn{2}{c}{$ 55.505_{\pm 0.066} $}                                                               & \multicolumn{2}{c}{$ 55.169_{\pm 0.066} $}                                                               & \multicolumn{2}{c}{$ 90.523_{\pm 0.009} $}                                                               \\ \cmidrule(l){3-11} 
                                                                                             & 135M                   & -                & \multicolumn{2}{c}{$ 19.242_{\pm 0.046} $}                                                               & \multicolumn{2}{c}{$ 42.686_{\pm 0.071} $}                                                               & \multicolumn{2}{c}{$ 43.664_{\pm 0.045} $}                                                               & \multicolumn{2}{c}{$ 90.448_{\pm 0.008} $}                                                               \\ \cmidrule(l){3-11} 
    \multirow{-2}{*}{\textbf{SmolLM2}}                                                       & 360M                   & -                & \multicolumn{2}{c}{$ 20.391_{\pm 0.027} $}                                                               & \multicolumn{2}{c}{$ 44.291_{\pm 0.095} $}                                                               & \multicolumn{2}{c}{$ 45.339_{\pm 0.062} $}                                                               & \multicolumn{2}{c}{$ 90.968_{\pm 0.010} $}                                                               \\ \cmidrule(l){3-11} 
    \textbf{Qwen2.5}                                                                         & 490M                   & -                & \multicolumn{2}{c}{$ 21.318_{\pm 0.045} $}                                                               & \multicolumn{2}{c}{$ 45.283_{\pm 0.081} $}                                                               & \multicolumn{2}{c}{$ 46.524_{\pm 0.169} $}                                                               & \multicolumn{2}{c}{$ 91.134_{\pm 0.014} $}                                                               \\ \midrule
    \multicolumn{11}{c}{\cellcolor[HTML]{FFFFFF}\textbf{Public Versions of Bible to American Standard Version of Bible (PUB-ASV)}}                                                                                                                                                                                                                                                                                                                                                                                                                                                   \\ \midrule
                                                                                             &                        & 8                & $ 28.770_{\pm 0.048} $                         & \cellcolor[HTML]{FFFFC7}$ 30.831_{\pm 0.065} $          & $ 56.341_{\pm 0.061} $                         & \cellcolor[HTML]{FFFFC7}$ 58.216_{\pm 0.047} $          & $ 51.586_{\pm 0.082} $                         & \cellcolor[HTML]{FFFFC7}$ 53.902_{\pm 0.038} $          & $ 89.543_{\pm 0.022} $                         & \cellcolor[HTML]{FFFFC7}$ 90.069_{\pm 0.040} $          \\
                                                                                             &                        & 16               & $ 39.887_{\pm 0.072} $                         & \cellcolor[HTML]{FFFFC7}$ 41.725_{\pm 0.035} $          & $ 64.846_{\pm 0.037} $                         & \cellcolor[HTML]{FFFFC7}$ 66.241_{\pm 0.035} $          & $ 62.210_{\pm 0.051} $                         & \cellcolor[HTML]{FFFFC7}$ 63.959_{\pm 0.049} $          & $ 92.047_{\pm 0.029} $                         & \cellcolor[HTML]{FFFFC7}$ 92.455_{\pm 0.029} $          \\
    \multirow{-3}{*}{\textbf{\begin{tabular}[c]{@{}c@{}}MDM w/o CFG\\ \end{tabular}}}  &                        & 64               & $ 47.610_{\pm 0.038} $                         & \cellcolor[HTML]{FFFFC7}$ 48.468_{\pm 0.035} $          & $ 70.085_{\pm 0.034} $                         & \cellcolor[HTML]{FFFFC7}$ 70.788_{\pm 0.023} $          & $ 68.469_{\pm 0.033} $                         & \cellcolor[HTML]{FFFFC7}$ 69.364_{\pm 0.022} $          & $ 93.627_{\pm 0.023} $                         & \cellcolor[HTML]{FFFFC7}$ 93.879_{\pm 0.036} $          \\ \cmidrule(l){3-11} 
                                                                                             &                        & 8                & \cellcolor[HTML]{FFCCC9}$ 33.535_{\pm 0.074} $ & \cellcolor[HTML]{FFCE93}$ 35.056_{\pm 0.041} $          & \cellcolor[HTML]{FFCCC9}$ 59.681_{\pm 0.053} $ & \cellcolor[HTML]{FFCE93}$ 61.051_{\pm 0.060} $          & \cellcolor[HTML]{FFCCC9}$ 56.314_{\pm 0.060} $ & \cellcolor[HTML]{FFCE93}$ 58.002_{\pm 0.077} $          & \cellcolor[HTML]{FFCCC9}$ 90.861_{\pm 0.027} $ & \cellcolor[HTML]{FFCE93}$ 91.246_{\pm 0.021} $          \\
                                                                                             &                        & 16               & \cellcolor[HTML]{FFCCC9}$ 43.060_{\pm 0.055} $ & \cellcolor[HTML]{FFCE93}$ 44.083_{\pm 0.042} $          & \cellcolor[HTML]{FFCCC9}$ 66.898_{\pm 0.042} $ & \cellcolor[HTML]{FFCE93}$ 67.709_{\pm 0.053} $          & \cellcolor[HTML]{FFCCC9}$ 65.378_{\pm 0.054} $ & \cellcolor[HTML]{FFCE93}$ 66.398_{\pm 0.054} $          & \cellcolor[HTML]{FFCCC9}$ 92.907_{\pm 0.022} $ & \cellcolor[HTML]{FFCE93}$ 93.124_{\pm 0.023} $          \\
    \multirow{-3}{*}{\textbf{\begin{tabular}[c]{@{}c@{}}MDM with CFG\\ \end{tabular}}} & \multirow{-6}{*}{162M} & 64               & \cellcolor[HTML]{FFCCC9}$ 50.695_{\pm 0.034} $ & \cellcolor[HTML]{FFCE93}$ \mathbf{51.118_{\pm 0.027}} $ & \cellcolor[HTML]{FFCCC9}$ 72.300_{\pm 0.021} $ & \cellcolor[HTML]{FFCE93}$ \mathbf{72.669_{\pm 0.023}} $ & \cellcolor[HTML]{FFCCC9}$ 71.906_{\pm 0.028} $ & \cellcolor[HTML]{FFCE93}$\mathbf{72.396_{\pm 0.028}} $  & \cellcolor[HTML]{FFCCC9}$ 94.516_{\pm 0.020} $ & \cellcolor[HTML]{FFCE93}$ \mathbf{94.608_{\pm 0.015}} $ \\ \cmidrule(l){3-11} 
    \textbf{GENIE}                                                                           & 144M                   & 64               & \multicolumn{2}{c}{$ 35.099_{\pm 0.060} $}                                                               & \multicolumn{2}{c}{$ 71.675_{\pm 0.033} $}                                                               & \multicolumn{2}{c}{$ 71.033_{\pm 0.050} $}                                                               & \multicolumn{2}{c}{$ 92.821_{\pm 0.008} $}                                                               \\ \cmidrule(l){3-11} 
                                                                                             & 135M                   & -                & \multicolumn{2}{c}{$ 33.620_{\pm 0.077} $}                                                               & \multicolumn{2}{c}{$ 57.254_{\pm 0.108} $}                                                               & \multicolumn{2}{c}{$ 57.138_{\pm 0.156} $}                                                               & \multicolumn{2}{c}{$ 92.310_{\pm 0.014} $}                                                               \\ \cmidrule(l){3-11} 
    \multirow{-2}{*}{\textbf{SmolLM2}}                                                       & 360M                   & -                & \multicolumn{2}{c}{$ 35.468_{\pm 0.072} $}                                                               & \multicolumn{2}{c}{$ 58.052_{\pm 0.072} $}                                                               & \multicolumn{2}{c}{$ 58.046_{\pm 0.097} $}                                                               & \multicolumn{2}{c}{$ 92.497_{\pm 0.015} $}                                                               \\ \cmidrule(l){3-11} 
    \textbf{Qwen2.5}                                                                         & 490M                   & -                & \multicolumn{2}{c}{$ 37.954_{\pm 0.154} $}                                                               & \multicolumn{2}{c}{$ 60.706_{\pm 0.037} $}                                                               & \multicolumn{2}{c}{$ 60.869_{\pm 0.022} $}                                                               & \multicolumn{2}{c}{$ 92.965_{\pm 0.005} $}                                                               \\ \bottomrule
    \end{tabular}%
    }
    \vspace{0.2cm}
    \caption{Main results for the Bible prose style transfer dataset. As mentioned in section \ref{subsec: metrics}, we report the mean and the standard deviation (in the subscript) across 8 independent sampling runs. $\uparrow$ indicates "higher the better" for the metric. The non-coloured cells report baseline scores, while the coloured cells represent different inference-time scaling settings.}
    \vspace{0.3cm}
    \label{table: bible}
\end{table*}

\subsection{Training and Inference Details}
For the experiments with MDM, we use pre-trained MDMs released by the work of \cite{smdm}. These MDMs are pre-trained on 627B tokens from the SlimPajama dataset \cite{cerebras2023slimpajama}. Specifically, we pick the MDM with 113M non-embedding parameters (167M total parameters) for fine-tuning. For the SVDD experiments, we use the \textit{MPNet-base-v2}\footnote{\url{https://huggingface.co/sentence-transformers/all-mpnet-base-v2}} sentence embedding model as a verifier for obtaining embedding cosine similarity rewards. For optimization, we use the AdamW optimizer \cite{adamw} with $ \beta_1 = 0.9 $ and $ \beta_2 = 0.95 $. We set a small value $ \epsilon = 10^{-5} $, indicating the minimum possible noise level \footnote{This is done to prevent numerical overflow in loss function calculation during training as for $ t \rightarrow 0 $, we will have $ \text{Loss} \rightarrow \infty $ given that we set $ \alpha_t = 1-t $ according to section \ref{sec: methodology}.}. To schedule the learning rate, we first implement a linear warmup schedule for a small number of warmup steps, after which we perform an inverse square root decay. All training experiments are done with \textit{bfloat16} precision except for the GENIE baseline, where \textit{fp32} precision is used. While training MDMs, we use an exponential moving average of weights during the weight update with a decay rate of 0.999. In the main results (table \ref{table: wikilarge} and \ref{table: bible}), the CFG results are reported with $ \gamma = 1.4$ and the SVDD results are reported with $M=4$ candidates per denoising step. The decoding with autoregressive LM baselines is performed with nucleus sampling with $ p=0.95 $. All of the experiments are performed on a single NVIDIA H100 GPU. Further details on hyperparameter settings for each experiment are provided in Appendix \ref{appendix C}.

\begin{figure*}[t]
\centering
    \begin{examplebox}[\small]
    
    \textbf{Input:} And David said to Saul, Let no man's heart fail because of him: thy servant will go and fight with this Philistine. \\ \\
    \textbf{Reference:} And David said to Saul, Let no man's heart become feeble because of him; I, your servant, will go out and have a fight with this Philistine. \\

    \begin{itemize}
        \setlength\itemsep{0.5em}
        \item \textbf{MDM ($\mathbf{M=4}$, $\mathbf{\gamma=1.4}$, steps = 64)}: And David said to Saul, Let no man's heart be feeble against him: your servant will go and make war on this Philistine.
        \item \textbf{MDM ($\mathbf{M=1}$, $\mathbf{\gamma=1.0}$, steps = 64)}: And David said to Saul, Let no man's heart be with him: your servant will go to war against this Philistine.
        \item \textbf{GENIE:} and david said to saul, let no man's heart be going because of him ; your servant will go and make war with the whole philistine.
        \item \textbf{SmolLM2-135M:} And David said to Saul, All men go to sleep because of him; let no man go out of his mind because of him: your servant will go in \textcolor{red}{flight when the Philistines are made strong.}

        \item \textbf{SmolLM2-360M:} And David said, Let no man be moved in heart by my words, because your servant will go and \textcolor{red}{put him to the sword.}

        \item \textbf{Qwen2.5:} And David said to Saul, Let no one be overcome because of him. Your servant will go and fight with this Philistine.
    \end{itemize}
    
    \vspace{4pt}
    \hrule
    \vspace{6pt}
    
    \textbf{Input:} And Gideon taketh ten men of his servants, and doth as Jehovah hath spoken unto him, and it cometh to pass, because he hath been afraid of the house of his father, and the men of the city, to do by day, that he doth by night. \\ \\
    \textbf{Reference:} Then Gideon took ten of his servants and did as the Lord had said to him; but fearing to do it by day, because of his father's people and the men of the town, he did it by night. \\

    \begin{itemize}
        \setlength\itemsep{0.5em}
        \item \textbf{MDM ($\mathbf{M=1}$, $\mathbf{\gamma=1.0}$, steps = 16)}: And Gideon took ten men of his servants and did, as the Lord had said to him, and because he \textcolor{blue}{was the fear of made his family} his \textcolor{blue}{and and} as whom the men of the year, so he did so by night

        \item \textbf{MDM ($\mathbf{M=4}$, $\mathbf{\gamma=1.0}$, steps = 16)}: And Gideon took ten men of his servants, did them \textcolor{blue}{as the said said}: and he was \textcolor{blue}{in fear and fear} of his family and because of of men of the town span the day \textcolor{blue}{he he night}

        \item \textbf{MDM ($\mathbf{M=4}$, $\mathbf{\gamma=1.4}$, steps = 16)}: So Gideon took ten men of his servants and did them as the Lord \textcolor{blue}{said said}: and he, because he was in \textcolor{blue}{fear fear} come of his father's family the who of the men of the town, he did so by night \textcolor{blue}{night night and night}.

        \item \textbf{MDM ($\mathbf{M=1}$, $\mathbf{\gamma=1.0}$, steps = 64)}: So Gideon took ten men of his servants and did as the Lord had said: and because he was fearing his father and his family and the men of David, \textcolor{blue}{days he did} it by night.

        \item \textbf{MDM ($\mathbf{M=4}$, $\mathbf{\gamma=1.0}$, steps = 64)}: So Gideon took ten men of his servants and did as the Lord had said: and because he was fearing them to his father and his men, he did them by night

        \item \textbf{MDM ($\mathbf{M=4}$, $\mathbf{\gamma=1.4}$, steps = 64)}: So Gideon took ten men of his servants and did as the Lord had said to him: and because he was full of fear of his father's family and the men of the town to do it by day, he did it by night.
    \end{itemize}
    \end{examplebox}
\caption{Examples of generated outputs from PUB-BBE task. The top example displays the outputs generated from the baseline methods and the MDM (with and without inference-time scaling). Out-of-context (and sometimes hallucinatory) behaviour in autoregressive LM generations is marked with \textcolor{red}{red}. The bottom example displays the outputs generated by the MDM with different inference-time hyperparameters. Lack of coherence in generated sentences is marked in \textcolor{blue}{blue}. More generated examples can be found in Figure \ref{fig: generated-examples-supp} of the supplementary material.}
\vspace{0.2cm}
\label{fig: generated-examples}
\end{figure*}

\subsection{Evaluation Metrics} \label{subsec: metrics}
We employ the following metrics for the evaluation of the generated outputs:

\begin{itemize}
    \item \textbf{BLEU \cite{bleu}: } A precision-based metric that measures the overlap between machine-generated text and one or more reference texts using n-grams.

    \item The Following metrics are used specifically for the evaluation of the WikiLarge simplicity style transfer dataset:

    \begin{itemize}
        \item \textbf{SARI \cite{sari}: }Evaluates the quality of sentence simplification by comparing system output against the source and reference texts, rewarding n-gram additions and deletions, and keeping operations that align with references.
        \item \textbf{LENS \cite{lens}: }It is a learned reference-based metric that scores simplifications based on fluency, adequacy, and simplicity.
    \end{itemize}

    \item The following metrics are used specifically for the evaluation of the Bible prose style transfer dataset:

    \begin{itemize}
        \item \textbf{ROUGE \cite{rouge}: } A recall-oriented set of metrics that evaluates the overlap of n-grams, word sequences, and word pairs between the candidate and reference texts.
        \item \textbf{METEOR \cite{meteor}: } Captures unigram matching between the model-generated text and the reference text, where unigrams can be matched based on their surface forms, stemmed forms, and meanings. 
        \item \textbf{BERTScore \cite{bertscore}: }Uses contextual embeddings from pre-trained BERT models to compute a similarity score between candidate and reference sentences, capturing semantic similarity better than n-gram-based metrics.
    \end{itemize}
\end{itemize}

During the evaluation phase, we perform multiple independent sampling runs on the test sets and report the mean and standard deviation of the metrics across the runs. For the WikiLarge dataset, we perform 20 runs, and for the Bible dataset, we perform 8 runs (due to a significantly larger test set than the former).

\section{Results and Discussion} \label{sec: results}
Table \ref{table: wikilarge} and Table \ref{table: bible} list out the main results for the WikiLarge dataset and the Bible dataset, respectively. As a general trend, we observe that the autoregressive baselines are outperformed by our diffusion baseline (i.e., GENIE). The diffusion baseline also outperforms the vanilla MDM setting (i.e., MDM without inference-time scaling). However, GENIE is pretrained with a case-insensitive tokenizer, which gives it a slight edge over other methods due to the lessened complexity of generation. This is specifically reflected in the scores for the Bible dataset, where we observe that the BLEU scores are significantly low for GENIE (as BLEU score computation involves direct n-gram matching) \footnote{Due to direct n-gram matching with the reference, if any capitalized letter occurs in a word, it won't match with the output of GENIE. For e.g., the word "Where" will not match with "where".}. Such a tokenizer design for generation can also be considered a downside since real-world scenarios would require ensuring proper capitalization during generation.

On the Bible test sets, the fine-tuned MDMs without any inference-time scaling beat the autoregressive baselines with as few as 16 steps. With the inclusion of classifier-free guidance and our proposed method with SVDD, we significantly outperform the diffusion model baseline as well. The first example of Figure \ref{fig: generated-examples} shows generated samples from the methods experimented with in this work. The autoregressive baselines tend to generate longer outputs and look less grounded in the input sentence. Oftentimes, we also observe out-of-context and hallucinatory behaviour from autoregressive LM generation (as highlighted in Figure \ref{fig: generated-examples}). Such behaviour is usually not seen in MDMs due to their non-autoregressive generation method. In contrast, MDM without inference-time scaling tends to generate short outputs, but in general produces more valid tokens\footnote{Additional results on the Bible dataset with static word vector embeddings can be found in Appendix \ref{appendix D}}.

On the Wikilarge test set, the fine-tuned MDMs without any inference-time scaling do not report promising simplicity scores. Qualitative analysis showed two reasons: \textbf{(1)} MDM produces very short outputs, possibly mistaking sentence shortening for simplicity, making the simplification inadequate in content, and \textbf{(2)} MDM tends to copy the input sentence almost exactly, resulting in no simplicity in the output. Both reasons lead to poor performance on simplicity metrics (specifically on LENS scores). However, such inconsistencies are vastly mitigated with inference-time scaling.

\subsection{Ablation Studies} \label{subsec: ablations}
\textbf{Classifier-free guidance (CFG): } Using classifier-free guidance during MDM decoding significantly improves the generation quality in both the TST tasks across different metrics (as shown in table \ref{table: wikilarge} and \ref{table: bible}). CFG improves the conditional grounding of the generated outputs using the guidance scale parameter. Intuitively, for $ \gamma > 1 $, we have a part of unconditional logits subtracted from the conditional logits (as per equation \ref{eqn: cfg}), which tends to improve the fidelity of sampled outputs from the resulting logits. Figure \ref{fig: plot-cfg} shows the effect of increasing the classifier-free guidance scale parameter. As the value increases, the performance also increases (due to an increasing influence of the conditional logits) up to a certain point, after which it either stagnates or slightly degrades. The degradation primarily happens due to over-reliance on the conditional signal, shifting the focus from the overall fluency of the generated sentence. \\

\noindent
\textbf{Soft-value diffusion decoding (SVDD): } Using SVDD with sentence embedding similarity as rewards consistently improves the generation performance of the MDM across different settings and metrics. The improvement is more significant with smaller denoising steps. The improvement can also be seen when using SVDD on top of CFG-based decoding. It is important to note that SVDD relies heavily on the inherent quality of the MDM itself. Intuitively, it tries to find the best solution amongst the pool of solutions that the MDM can generate. SVDD with our proposed reward aims to achieve this by performing semantic content maximization between the input and the generated sentences along the denoising trajectory at every timestep. Increasing the number of possible candidates for verification leads to better final estimates. 

\subsection{Test-time compute scaling} \label{sec: test-time compute}
 The bottom example of Figure \ref{fig: generated-examples} shows the effect on the generation with different MDM inference settings. MDM generation without inference-time scaling often struggles with sentence coherence and tends to produce unnecessary repeated tokens. Below, we discuss in more detail the effect of scaling the finetuned MDMs along different scaling axes, where diffusion models offer the flexibility to do so.  \\

\begin{figure}[t]
    \centering
    \begin{subfigure}{0.47\textwidth}
        \includegraphics[width=\linewidth]{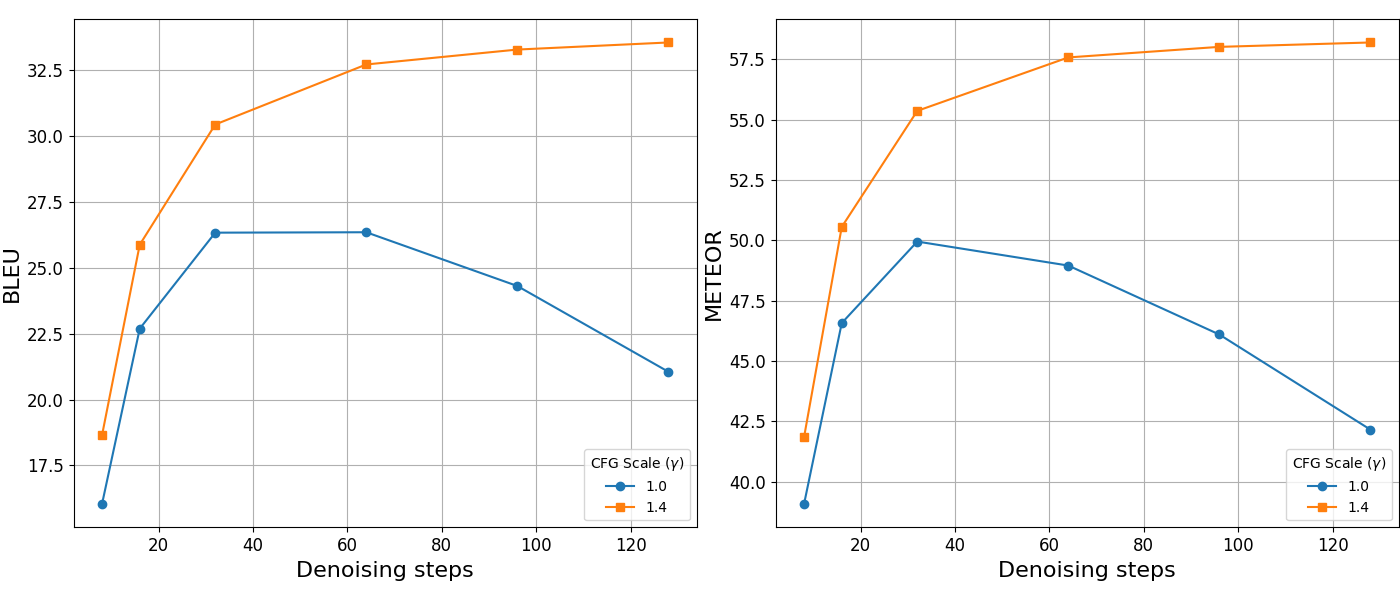}
        \vspace{-25pt}
        \caption{}
        \label{fig: plot-step-scaling} 
    \end{subfigure}%
    \vspace{0.5cm} 
    \begin{subfigure}{0.47\textwidth}
        \includegraphics[width=\linewidth]{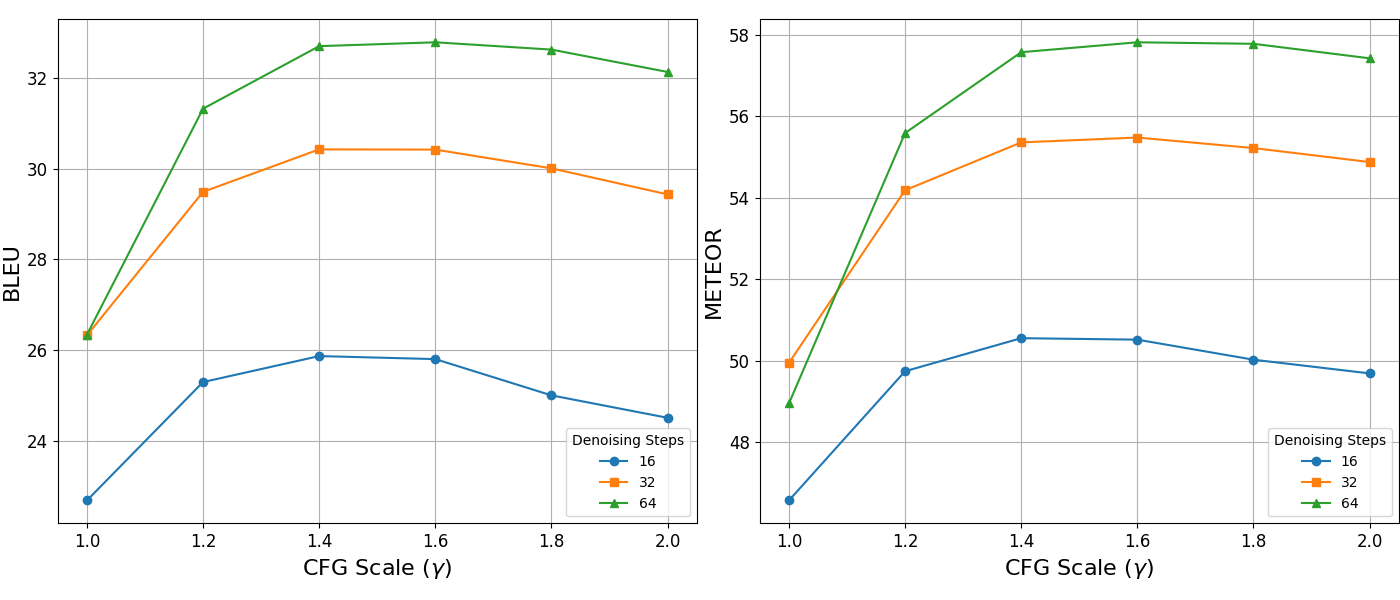}
        \vspace{-25pt}
        \caption{}
        \label{fig: plot-cfg} 
    \end{subfigure}%
    \vspace{0.5cm} 
    \begin{subfigure}{0.47\textwidth}
        \includegraphics[width=\linewidth]{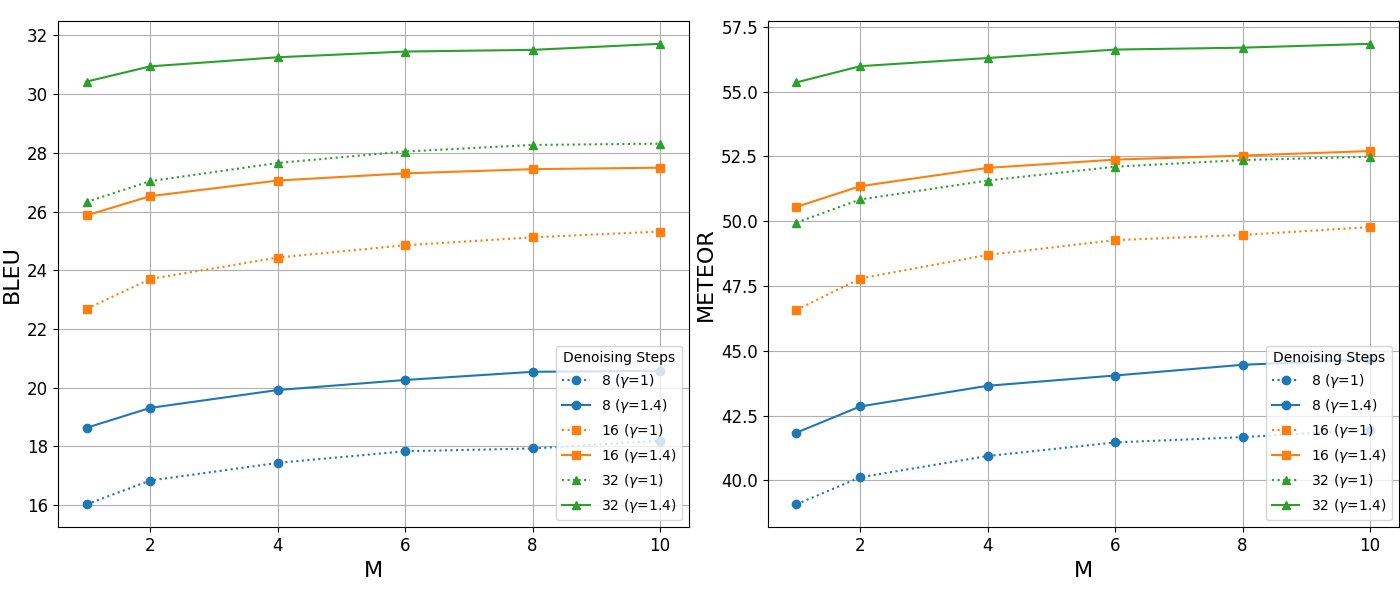}
        \vspace{-25pt}
        \caption{}
        \label{fig: plot-M} 
    \end{subfigure}
    \vspace{0.5cm}
    \caption{Plots for performance metrics vs. inference-time hyperparameters as evaluated on the PUB-BBE test set. $ \gamma $ is the CFG scale. \textbf{(a)} Shows the effect of scaling across the timestep axis with CFG (i.e., $\gamma = 1.4$) and without CFG (i.e., $ \gamma = 1 $), \textbf{(b)} shows the effect of the CFG scale ($\gamma$) during inference, and \textbf{(c)} shows the effect of scaling across the verifier axis by increasing the number of candidates (i.e., $M$) to be generated each step for verification. Refer to Appendix \ref{appendix E} for plots on the other two test sets.}
    \vspace{0.8cm}
    \label{fig: plots} 
\end{figure}

\textbf{Effect of scaling denoising timesteps: } Figure \ref{fig: plot-step-scaling} shows the plots for metrics vs. denoising steps. In general, increasing denoising timesteps (keeping other hyperparameters the same) leads to improving performance up to a certain point, after which it stagnates \cite{compute_scaling_diffusion}. However, vanilla MDM decoding observes a slight drop in performance after moving above a certain number of denoising timesteps. This observation is slightly more pronounced in the simplicity style transfer task, where the MDM starts generating shorter outputs with more denoising steps. While shorter sentences are simpler, in a conditional setting, they often fail to capture the content, resulting in a drop in performance. The addition of classifier-free guidance allows MDMs to scale better with increasing time steps, showing trends similar to diffusion timestep scaling for images. \\

\textbf{Effect of number of candidates in an SVDD step: } Figure \ref{fig: plot-M} shows the effect of increasing the number of candidates verified per denoising step on the performance metrics. Similar to scaling denoising steps, increasing the number of candidates also stagnates after a certain number. Having a fixed number of candidates to be considered for verification is like a Monte-Carlo approximation for choosing the best among all possible candidates at a particular step. This means a larger number of candidates will lead to a better approximation. However, the number of forward passes to be made per step linearly increases with the number of candidates, making it essential to consider this compute-time vs. quality tradeoff carefully. The stagnation in performance after a certain number of candidates indicates that the current pool of candidates is nearly sufficient to represent the pool of all possible candidates. We also observe that SVDD improves more upon MDM inference without CFG compared to inference with CFG. Both stagnation with increasing candidates and the effect of CFG can be attributed to the diversity that the MDM sampling can exhibit. The number of verification candidates can potentially better scale the performance if the model exhibits high generation diversity, as there would be more possible solutions. Since CFG improves generation quality by sacrificing diversity, it also explains why SVDD would be slightly less effective during inference with CFG, as mentioned earlier. 

\section{Related Work}
\textbf{Text style transfer (TST)} has been a widely explored NLP task consisting of many dimensions like simplicity, formality, prose, sentiment, etc. Over the years, much work has gone into these tasks, considering both supervised and unsupervised settings. Apart from the traditional seq-to-seq setups, several multi-task learning approaches have been proposed for supervised settings \cite{multitask-tst}. Also, synthetic data generation is usually performed where parallel data is scarce. For unsupervised settings, distentangled representation learning for the style and content of the text is the most popular approach \cite{li2020complementary}. The survey of \cite{tst-survey} gives a comprehensive list of methods in both categories. \\

\noindent
\textbf{Discrete diffusion models} are discrete variants of the popular diffusion modeling framework used for generating images. The framework aimed to apply diffusion modeling to discrete data modalities like text, graphs, protein sequences, etc. Due to their simplicity, two common variants widely studied for discrete diffusion are the uniform and the absorbing (or masked) variants \cite{multinomial_diffusion, austin_diffusion, sedd}. Amongst these two, masked discrete diffusion has proven to be a more scalable approach and has been competitive with autoregressive LMs \cite{mdlm, smdm, sedd}. Scaling diffusion models during inference with search algorithms and external verifiers \cite{compute_scaling_diffusion} has allowed diffusion-based generation to scale beyond just increasing denoising steps (which is known to quickly stagnate). These techniques include a vast literature of diffusion guidance algorithms \cite{classifier_guidance, cfg, tfg} that enable controllability during generation. Recently, guidance mechanisms have also been extended to discrete diffusion settings \cite{simple_discrete_guidance, unlocking_discrete_guidance}, opening potential research directions in scaling discrete diffusion language models during inference.  

The work of \cite{diff_style_transfer} explores supervised TST using seq-to-seq continuous diffusion models \cite{diffuseq}. While this is the only work combining diffusion models and TST (to the best of our knowledge), the work limits itself by only showcasing the application of diffusion models for TST, leaving many capabilities of diffusion models unexplored.   

\section{Summary, Conclusion and Future Work}
In this paper, we present masked diffusion language models (MDMs) for the task of text style transfer. We explore verifier-based inference-time scaling in MDMs and propose pre-trained sentence embedding models as verifiers. This significantly improves generated outputs by combining the proposed verifier with soft-value diffusion decoding. Additionally, we analyze the scaling behaviour of MDM inference along different scaling axes. Our method is simple and does not require additional training on top of fine-tuning the MDMs since we can directly incorporate pre-trained embedding models as verifiers. Our work not only establishes MDMs as a better alternative for TST but also emphasizes the benefits of inference-time scaling in diffusion language models. Our work can be expanded into two broad directions: \textbf{(1)} train external verifiers for task-specific use cases, potentially resulting in better outputs at the cost of extra training, and \textbf{(2)} expand the methodology for different downstream tasks by designing task-specific reward functions. We leave further exploration in these two directions for future work. 

\bibliography{mybibfile}

\begin{thebibliography}{39}
\providecommand{\natexlab}[1]{#1}
\providecommand{\url}[1]{\texttt{#1}}
\expandafter\ifx\csname urlstyle\endcsname\relax
  \providecommand{\doi}[1]{doi: #1}\else
  \providecommand{\doi}{doi: \begingroup \urlstyle{rm}\Url}\fi

\bibitem[Allal and et~al.(2025)]{smollm2}
L.~B. Allal and et~al.
\newblock Smollm2: When smol goes big - data-centric training of a small language model.
\newblock \emph{CoRR}, abs/2502.02737, 2025.

\bibitem[Austin et~al.(2021)Austin, Johnson, Ho, Tarlow, and van~den Berg]{austin_diffusion}
J.~Austin, D.~D. Johnson, J.~Ho, D.~Tarlow, and R.~van~den Berg.
\newblock Structured denoising diffusion models in discrete state-spaces.
\newblock In \emph{Proceedings of the 35th International Conference on Neural Information Processing Systems}, NIPS '21, Red Hook, NY, USA, 2021. Curran Associates Inc.
\newblock ISBN 9781713845393.

\bibitem[Banerjee and Lavie(2005)]{meteor}
S.~Banerjee and A.~Lavie.
\newblock {METEOR}: An automatic metric for {MT} evaluation with improved correlation with human judgments.
\newblock In \emph{Proceedings of the {ACL} Workshop on Intrinsic and Extrinsic Evaluation Measures for Machine Translation and/or Summarization}, pages 65--72, Ann Arbor, Michigan, June 2005.

\bibitem[Brown and et~al.(2020)]{gpt3}
T.~B. Brown and et~al.
\newblock Language models are few-shot learners.
\newblock In \emph{Proceedings of the 34th International Conference on Neural Information Processing Systems}, NIPS '20, Red Hook, NY, USA, 2020. Curran Associates Inc.
\newblock ISBN 9781713829546.

\bibitem[Carlson et~al.(2017)Carlson, Riddell, and Rockmore]{bible}
K.~Carlson, A.~B. Riddell, and D.~N. Rockmore.
\newblock Evaluating prose style transfer with the bible.
\newblock \emph{Royal Society Open Science}, 5, 2017.
\newblock URL \url{https://api.semanticscholar.org/CorpusID:21186239}.

\bibitem[Chang et~al.(2022)Chang, Zhang, Jiang, Liu, and Freeman]{maskgit}
H.~Chang, H.~Zhang, L.~Jiang, C.~Liu, and W.~T. Freeman.
\newblock Maskgit: Masked generative image transformer.
\newblock \emph{2022 IEEE/CVF Conference on Computer Vision and Pattern Recognition (CVPR)}, pages 11305--11315, 2022.

\bibitem[Devlin et~al.(2019)Devlin, Chang, Lee, and Toutanova]{bert}
J.~Devlin, M.-W. Chang, K.~Lee, and K.~Toutanova.
\newblock {BERT}: Pre-training of deep bidirectional transformers for language understanding.
\newblock In \emph{Proceedings of the 2019 Conference of the North {A}merican Chapter of the Association for Computational Linguistics: Human Language Technologies, Volume 1 (Long and Short Papers)}, pages 4171--4186, Minneapolis, Minnesota, June 2019.

\bibitem[Dhariwal and Nichol(2021)]{classifier_guidance}
P.~Dhariwal and A.~Nichol.
\newblock Diffusion models beat gans on image synthesis.
\newblock In \emph{Proceedings of the 35th International Conference on Neural Information Processing Systems}, NIPS '21, Red Hook, NY, USA, 2021. Curran Associates Inc.
\newblock ISBN 9781713845393.

\bibitem[Dou and Song(2024)]{dps-smc}
Z.~Dou and Y.~Song.
\newblock Diffusion posterior sampling for linear inverse problem solving: {A} filtering perspective.
\newblock In \emph{The Twelfth International Conference on Learning Representations, {ICLR} 2024, Vienna, Austria, May 7-11, 2024}. OpenReview.net, 2024.

\bibitem[Dubey and et~al.(2024)]{llama3}
A.~Dubey and et~al.
\newblock The llama 3 herd of models.
\newblock \emph{CoRR}, abs/2407.21783, 2024.

\bibitem[Gong et~al.(2023)Gong, Li, Feng, Wu, and Kong]{diffuseq}
S.~Gong, M.~Li, J.~Feng, Z.~Wu, and L.~Kong.
\newblock Diffuseq: Sequence to sequence text generation with diffusion models.
\newblock In \emph{The Eleventh International Conference on Learning Representations, {ICLR} 2023, Kigali, Rwanda, May 1-5, 2023}. OpenReview.net, 2023.

\bibitem[Ho and Salimans(2021)]{cfg}
J.~Ho and T.~Salimans.
\newblock Classifier-free diffusion guidance.
\newblock In \emph{NeurIPS 2021 Workshop on Deep Generative Models and Downstream Applications}, 2021.
\newblock URL \url{https://openreview.net/forum?id=qw8AKxfYbI}.

\bibitem[Hoogeboom et~al.(2021)Hoogeboom, Nielsen, Jaini, Forr'e, and Welling]{multinomial_diffusion}
E.~Hoogeboom, D.~Nielsen, P.~Jaini, P.~Forr'e, and M.~Welling.
\newblock Argmax flows and multinomial diffusion: Learning categorical distributions.
\newblock In \emph{Neural Information Processing Systems}, 2021.

\bibitem[Jin et~al.(2022)Jin, Jin, Hu, Vechtomova, and Mihalcea]{tst-survey}
D.~Jin, Z.~Jin, Z.~Hu, O.~Vechtomova, and R.~Mihalcea.
\newblock Deep learning for text style transfer: A survey.
\newblock \emph{Computational Linguistics}, 48\penalty0 (1):\penalty0 155--205, Mar. 2022.

\bibitem[Lee et~al.(2025)Lee, Jeha, Frellsen, Lio, Albergo, and Vargas]{debiasing_guidance}
C.~K. Lee, P.~Jeha, J.~Frellsen, P.~Lio, M.~S. Albergo, and F.~Vargas.
\newblock Debiasing guidance for discrete diffusion with sequential monte carlo.
\newblock \emph{CoRR}, abs/2502.06079, 2025.

\bibitem[Li et~al.(2024)Li, Zhao, Wang, Scalia, Eraslan, Nair, Biancalani, Regev, Levine, and Uehara]{svdd}
X.~Li, Y.~Zhao, C.~Wang, G.~Scalia, G.~Eraslan, S.~Nair, T.~Biancalani, A.~Regev, S.~Levine, and M.~Uehara.
\newblock Derivative-free guidance in continuous and discrete diffusion models with soft value-based decoding.
\newblock \emph{CoRR}, abs/2408.08252, 2024.

\bibitem[Li et~al.(2020)Li, Li, Zhang, Li, Zheng, Carin, and Gao]{li2020complementary}
Y.~Li, C.~Li, Y.~Zhang, X.~Li, G.~Zheng, L.~Carin, and J.~Gao.
\newblock Complementary auxiliary classifiers for label-conditional text generation.
\newblock In \emph{Proceedings of the AAAI Conference on Artificial Intelligence}, volume~34, pages 8303--8310, 2020.

\bibitem[Lin(2004)]{rouge}
C.-Y. Lin.
\newblock {ROUGE}: A package for automatic evaluation of summaries.
\newblock In \emph{Text Summarization Branches Out}, pages 74--81, Barcelona, Spain, July 2004. Association for Computational Linguistics.

\bibitem[Lin et~al.(2023)Lin, Gong, Shen, Wu, Fan, Lin, Duan, and Chen]{GENIE}
Z.~Lin, Y.~Gong, Y.~Shen, T.~Wu, Z.~Fan, C.~Lin, N.~Duan, and W.~Chen.
\newblock Text generation with diffusion language models: a pre-training approach with continuous paragraph denoise.
\newblock In \emph{Proceedings of the 40th International Conference on Machine Learning}, ICML'23. JMLR.org, 2023.

\bibitem[Loshchilov and Hutter(2019)]{adamw}
I.~Loshchilov and F.~Hutter.
\newblock Decoupled weight decay regularization.
\newblock In \emph{7th International Conference on Learning Representations, {ICLR} 2019, New Orleans, LA, USA, May 6-9, 2019}. OpenReview.net, 2019.

\bibitem[Lou et~al.(2024)Lou, Meng, and Ermon]{sedd}
A.~Lou, C.~Meng, and S.~Ermon.
\newblock Discrete diffusion modeling by estimating the ratios of the data distribution.
\newblock In \emph{Forty-first International Conference on Machine Learning, {ICML} 2024, Vienna, Austria, July 21-27, 2024}. OpenReview.net, 2024.

\bibitem[Lyu et~al.(2023)Lyu, Luo, Shi, Hollon, and Lee]{diff_style_transfer}
Y.~Lyu, T.~Luo, J.~Shi, T.~Hollon, and H.~Lee.
\newblock Fine-grained text style transfer with diffusion-based language models.
\newblock In \emph{Proceedings of the 8th Workshop on Representation Learning for NLP (RepL4NLP 2023)}, pages 65--74, Toronto, Canada, July 2023.

\bibitem[Ma et~al.(2025)Ma, Tong, Jia, Hu, Su, Zhang, Yang, Li, Jaakkola, Jia, and Xie]{compute_scaling_diffusion}
N.~Ma, S.~Tong, H.~Jia, H.~Hu, Y.~Su, M.~Zhang, X.~Yang, Y.~Li, T.~S. Jaakkola, X.~Jia, and S.~Xie.
\newblock Inference-time scaling for diffusion models beyond scaling denoising steps.
\newblock \emph{CoRR}, abs/2501.09732, 2025.
\newblock \doi{10.48550/ARXIV.2501.09732}.
\newblock URL \url{https://doi.org/10.48550/arXiv.2501.09732}.

\bibitem[Maddela et~al.(2023)Maddela, Dou, Heineman, and Xu]{lens}
M.~Maddela, Y.~Dou, D.~Heineman, and W.~Xu.
\newblock {LENS}: A learnable evaluation metric for text simplification.
\newblock In \emph{Proceedings of the 61st Annual Meeting of the Association for Computational Linguistics (Volume 1: Long Papers)}, pages 16383--16408, Toronto, Canada, July 2023.

\bibitem[Nie et~al.(2025{\natexlab{a}})Nie, Zhu, Du, Pang, Liu, Zeng, Lin, and Li]{smdm}
S.~Nie, F.~Zhu, C.~Du, T.~Pang, Q.~Liu, G.~Zeng, M.~Lin, and C.~Li.
\newblock Scaling up masked diffusion models on text.
\newblock In \emph{The Thirteenth International Conference on Learning Representations}, 2025{\natexlab{a}}.
\newblock URL \url{https://openreview.net/forum?id=WNvvwK0tut}.

\bibitem[Nie et~al.(2025{\natexlab{b}})Nie, Zhu, You, Zhang, Ou, Hu, Zhou, Lin, Wen, and Li]{llada}
S.~Nie, F.~Zhu, Z.~You, X.~Zhang, J.~Ou, J.~Hu, J.~Zhou, Y.~Lin, J.~Wen, and C.~Li.
\newblock Large language diffusion models.
\newblock \emph{CoRR}, abs/2502.09992, 2025{\natexlab{b}}.
\newblock \doi{10.48550/ARXIV.2502.09992}.
\newblock URL \url{https://doi.org/10.48550/arXiv.2502.09992}.

\bibitem[Nisonoff et~al.(2024)Nisonoff, Xiong, Allenspach, and Listgarten]{unlocking_discrete_guidance}
H.~Nisonoff, J.~Xiong, S.~Allenspach, and J.~Listgarten.
\newblock Unlocking guidance for discrete state-space diffusion and flow models.
\newblock \emph{CoRR}, abs/2406.01572, 2024.
\newblock \doi{10.48550/ARXIV.2406.01572}.
\newblock URL \url{https://doi.org/10.48550/arXiv.2406.01572}.

\bibitem[Niu et~al.(2018)Niu, Rao, and Carpuat]{multitask-tst}
X.~Niu, S.~Rao, and M.~Carpuat.
\newblock Multi-task neural models for translating between styles within and across languages.
\newblock In E.~M. Bender, L.~Derczynski, and P.~Isabelle, editors, \emph{Proceedings of the 27th International Conference on Computational Linguistics}, pages 1008--1021, Aug. 2018.

\bibitem[Papineni et~al.(2002)Papineni, Roukos, Ward, and Zhu]{bleu}
K.~Papineni, S.~Roukos, T.~Ward, and W.-J. Zhu.
\newblock {B}leu: a method for automatic evaluation of machine translation.
\newblock In P.~Isabelle, E.~Charniak, and D.~Lin, editors, \emph{Proceedings of the 40th Annual Meeting of the Association for Computational Linguistics}, pages 311--318, Philadelphia, Pennsylvania, USA, July 2002.

\bibitem[Sahoo et~al.(2024)Sahoo, Arriola, Schiff, Gokaslan, Marroquin, Chiu, Rush, and Kuleshov]{mdlm}
S.~S. Sahoo, M.~Arriola, Y.~Schiff, A.~Gokaslan, E.~Marroquin, J.~T. Chiu, A.~Rush, and V.~Kuleshov.
\newblock Simple and effective masked diffusion language models.
\newblock In \emph{Annual Conference on Neural Information Processing Systems 2024, NeurIPS 2024, Vancouver, BC, Canada, December 10 - 15, 2024}, 2024.

\bibitem[Schiff et~al.(2025)Schiff, Sahoo, Phung, Wang, Boshar, Dalla-torre, de~Almeida, Rush, PIERROT, and Kuleshov]{simple_discrete_guidance}
Y.~Schiff, S.~S. Sahoo, H.~Phung, G.~Wang, S.~Boshar, H.~Dalla-torre, B.~P. de~Almeida, A.~M. Rush, T.~PIERROT, and V.~Kuleshov.
\newblock Simple guidance mechanisms for discrete diffusion models.
\newblock In \emph{The Thirteenth International Conference on Learning Representations}, 2025.
\newblock URL \url{https://openreview.net/forum?id=i5MrJ6g5G1}.

\bibitem[Shi et~al.(2024)Shi, Han, Wang, Doucet, and Titsias]{shi2024simplified}
J.~Shi, K.~Han, Z.~Wang, A.~Doucet, and M.~Titsias.
\newblock Simplified and generalized masked diffusion for discrete data.
\newblock In \emph{The Thirty-eighth Annual Conference on Neural Information Processing Systems}, 2024.

\bibitem[Soboleva et~al.(2023)Soboleva, Al-Khateeb, Myers, Steeves, Hestness, and Dey]{cerebras2023slimpajama}
D.~Soboleva, F.~Al-Khateeb, R.~Myers, J.~R. Steeves, J.~Hestness, and N.~Dey.
\newblock {SlimPajama: A 627B token cleaned and deduplicated version of RedPajama}, 2023.
\newblock URL \url{https://huggingface.co/datasets/cerebras/SlimPajama-627B}.

\bibitem[Song et~al.(2021)Song, Meng, and Ermon]{ddim}
J.~Song, C.~Meng, and S.~Ermon.
\newblock Denoising diffusion implicit models.
\newblock In \emph{International Conference on Learning Representations}, 2021.

\bibitem[Xu et~al.(2016)Xu, Napoles, Pavlick, Chen, and Callison-Burch]{sari}
W.~Xu, C.~Napoles, E.~Pavlick, Q.~Chen, and C.~Callison-Burch.
\newblock Optimizing statistical machine translation for text simplification.
\newblock \emph{Transactions of the Association for Computational Linguistics}, 4:\penalty0 401--415, 2016.

\bibitem[Yang and et~al.(2024)]{qwen2.5}
A.~Yang and et~al.
\newblock Qwen2.5 technical report.
\newblock \emph{CoRR}, abs/2412.15115, 2024.
\newblock \doi{10.48550/ARXIV.2412.15115}.
\newblock URL \url{https://doi.org/10.48550/arXiv.2412.15115}.

\bibitem[Ye and et~al.(2024)]{tfg}
H.~Ye and et~al.
\newblock {TFG}: Unified training-free guidance for diffusion models.
\newblock In \emph{The Thirty-eighth Annual Conference on Neural Information Processing Systems}, 2024.

\bibitem[Zhang et~al.(2020)Zhang, Kishore, Wu, Weinberger, and Artzi]{bertscore}
T.~Zhang, V.~Kishore, F.~Wu, K.~Q. Weinberger, and Y.~Artzi.
\newblock Bertscore: Evaluating text generation with {BERT}.
\newblock In \emph{8th International Conference on Learning Representations, {ICLR} 2020, Addis Ababa, Ethiopia, April 26-30, 2020}. OpenReview.net, 2020.

\bibitem[Zhang and Lapata()]{wikilarge}
X.~Zhang and M.~Lapata.
\newblock Sentence simplification with deep reinforcement learning.
\newblock In \emph{Proceedings of the 2017 Conference on Empirical Methods in Natural Language Processing}, pages 584--594.

\end{thebibliography}

\newpage
\appendix
\section*{Appendices}
\section{Training and Sampling Algorithms} \label{appendix A}

Algorithm \ref{alg: training} gives the training algorithm for MDMs. For each training sample $x_0$, we sample a random $t$, run the forward masked diffusion process, and compute the loss function (i.e., the cross-entropy loss on the masked positions). After loss function computation, we take the optimization step by computing the loss gradients.
\begin{algorithm}[h]
    \caption{MDM Training Algorithm} \label{alg: svdd}
    
    \textbf{Given:} Denoiser $ \text{NN}_{\theta} $, input/source sentence $y$, mask token $\mtok$.  \\
    \Repeat{convergence}{
        \textbf{Forward process: } $x_0 \sim $ data, $ t \sim \mathcal{U}(0,1) $, and $ x_t \sim q_{t|0}(x_t | x_0) $ \\
        \textbf{Compute:} $ p_{\theta}( x_0^i|x_t, y ) = \text{NN}_{\theta}(x_t, y) $ \\
        \textbf{Compute loss:} $\mathcal{L}_{\text{MDM}} = -\underset{x_t^i = \mtok}{\sum} \log p_{\theta}( x_0^i|x_t, y )$ \\
        Gradient descent with $ \nabla_{\theta} \mathcal{L}_{\text{MDM}} $
    }\label{alg: training}
\end{algorithm}

Algorithm \ref{alg: sampling} gives the sampling algorithm for MDMs. In our work, we use the greedy decoding algorithm proposed by \cite{maskgit}. Let $ k $ be the number of tokens to be unmasked at an arbitrary timestep. The algorithm then chooses the $top-k$ tokens predicted with the highest probability using $ p_{\theta}(x_0 | x_t) $. The prediction $ p_{\theta}(x_0 | x_t) $ can also be done using classifier-free guidance as described in Section 3.1 of the main content.

\begin{algorithm}[h]
    \caption{MDM Sampling Algorithm} \label{alg: svdd}
    
    \textbf{Given:} Denoiser $ \text{NN}_{\theta} $, input/source sentence $y$, Total denoising steps = $T$, generation sequence length $L$, mask token $\mtok$.  \\
    \textbf{Define:} $c_i$ as the probability of a token $ x_0^i $ as assigned by the distribution $ p_{\theta}(x_0^i | x_t) $. \\
    \textbf{Initialize} $x_T$ as a sequence of all masked tokens of length $L$. \\
    \For{$t = 1, \frac{T-1}{T}, \frac{T-2}{T}, ..., \frac{1}{T}$}{
        $ s = t - \frac{1}{T} $ \\
        $k = \lfloor L(1-s) \rfloor$ \\
        $ p_{\theta}( x_0^i|x_t, y ) = \text{NN}_{\theta}(x_t, y) $ \\
        \For{$i= [1,2, ..., L]$}{
            \eIf{$ x_t^i \neq \mtok $}{
                $x_0^i = x_t^i \text{ and } c_i = 1$ \\
            }{
                $ x_0^i = \text{argmax}_j \,p_{\theta}(x_0^i | x_t, y) \text{ and } c_i = p_{\theta}(x_0^i|x_t, y)_{x_0^i}$ \\
            }
            \If{$ c_i \in \text{top-}k \left( \{ c_l \}_{i=1}^{L-1} \right) $}{
                $ x_s^i = x_0^i$ \\
            }
        }    
    }
    return $ x_0 $
    \label{alg: sampling}
\end{algorithm}

The SVDD algorithm using the proposed sentence-embedding verifier can be easily incorporated into the algorithm by plugging in the candidate selection process into the part where we compute $ x_s$  from $ x_t$. Specifically, at each denoising step, we obtain multiple samples for $ x_s $ and follow the SVDD algorithm (described in Section 3.2 and Algorithm 1 of the main content) to select the best $ x_s $ candidate for that denoising step. 

\section{Soft-Value Diffusion Decoding} \label{appendix B}
Soft-value diffusion decoding is based on the principle of generation via reward maximization while preserving the original distribution that was initially trained for generating a particular distribution. Let $ p_{\theta}(\cdot) $ be the approximate data distribution induced by a pre-trained neural network with parameters $ \theta $. Formally, we aim to sample from a new distribution $ p_{\alpha}(\cdot) $ which is defined as

\begin{align}
    p_{\alpha}(x) :&= \underset{p}{\text{argmax}} \left [ \mathbb{E}_{x\sim p(\cdot)}\left( r(x) \right) - \alpha \, \text{KL}\left( p(\cdot) || p_{\theta}(\cdot) \right) \right]. 
\end{align}

Here, the first term corresponds to reward maximization where $r(\cdot)$ is the reward function, and the second term aims to preserve the pretrained knowledge about the distribution learned by $ p_{\theta}(\cdot) $. The definition can be reduced as follows: 

\begin{align}
    p_{\alpha}(x) &= \underset{p}{\text{argmax}} \left [ \mathbb{E}_{x\sim p(\cdot)}\left( \alpha \log e^{r(x)/\alpha} \right) - \right. \notag \\
    &\hspace{2.8cm}\left. \mathbb{E}_{x\sim p(\cdot)} \left( \alpha \log \frac{p(x)}{p_{\theta}(x)} \right)  \right], \notag \\
    &\propto \underset{p}{\text{argmax}} \left[ e^{r(x)/\alpha} p_{\theta}(x) \right].& 
    \label{eqn: propto}
\end{align}

We define a value function from the perspective of Masked Diffusion Language Models and show how soft-value diffusion decoding approximates the optimal denoising trajectory. The value function $ v_{t}(\cdot) $ gives the expected reward (in the future after the reverse process is finished) at $ t=0 $ for an arbitrary noisy state $ t $. Based on equation \ref{eqn: propto}, we formally write

\begin{align}
    v_{t}(\cdot) := \alpha  \log \mathbb{E}_{x_0 \sim p_{\theta}(x_0 | x_t)} \left[ e^{r(x_0) / \alpha} | x_{t} = \cdot \right].
\end{align}

Here, $ \mathbb{E}_{x_0 \sim p_{\theta}(x_0 | x_t)}(\cdot) $ is induced by $ p_{\theta}(x_{t} | x_{t+1}) $. Assuming a time discretization of $t \in \{T+1, T, ..., t, t-1, ..., 1\}$ in the diffusion sampling process, we can write $ p_{\alpha}(x_0) = \int \prod_{t=T}^0 p_{\alpha}(x_t | x_{t+1}) \, dx_{1:T} $. The aim is to sample from the denoising trajectory for reward maximization. To do this, \cite{svdd} performs importance sampling at every denoising step as follows:

\begin{align}
    p_{\alpha}^{\star}(x_t | x_{t+1}) &\approx \sum_{m=1}^{M} \frac{\omega_{t}^{(m)}}{\sum_{i=1}^{M} \omega_{t}^{(i)} } \, \delta(x_{t}^{(m)}), \notag \\
    \text{where } \; \{x_{t}^{m}\}_{m=1}^{M} &\sim p_{\theta}(x_t | x_{t+1}) \text{ and } \omega_t^{(m)} := e^{v_t\left(x_t^{(m)}\right) / \alpha}.
\end{align}

Here, $M$ is the number of candidate samples considered for $ x_t$ to approximate the reward-maximizing denoising trajectory, and $ \delta(\cdot)$ is the Kronecker delta function. The importance sampling weights $\omega_t$ are defined in terms of the value function (i.e., the expected reward). This allows us to steer the denoising trajectory towards generation with higher rewards. In other words, sampling $x_t$ from $ x_{t+1}$ is broken down into two steps: \textbf{(1)} sample $M$ $x_t$ candidates and \textbf{(2)} sample an index $ \xi \in [1,2, ..., M] $ with $[\omega_t^{(1)}, \omega_t^{(2)}, ..., \omega_t^{(M)}]$ as unnormalized probability mass values thereby giving us $ x_t^{(\xi)} $ as the final selected candidate. 

The value function can be approximated using the posterior-mean-approximation (PMA) method \cite{svdd}. This method defines a reward function on $ x_0$ (i.e., the clean data estimate) and then computes the value function $ v_t(\cdot) $ by first sampling an $x_0$ estimate from $ p_{\theta}(x_0 | x_t) $ and then passing the estimate to the reward function.

\begin{align}
    v_t(x_t) &:= \alpha  \log \mathbb{E}_{x_0 \sim p_{\theta}(x_0 | x_t)} \left[ e^{r(x_0) / \alpha} | x_{t} \right] \notag \\
    & \approx \alpha \log \left( e^{r \left( \widehat{x}_0(x_t) \right) / \alpha} \right) \notag \\
    &= r \left( \widehat{x}_0(x_t) \right).
\end{align}

Here, $ \widehat{x}_0(x_t) \sim p_{\theta}(x_0 | x_t) $ is used to approximate the value function. Since the reward function takes an estimate of clean data, it is possible to use a pre-trained model as a reward function directly without extra training.

\begin{figure}
    \centering
    \begin{subfigure}[b]{0.45\textwidth}
        \includegraphics[width=\textwidth]{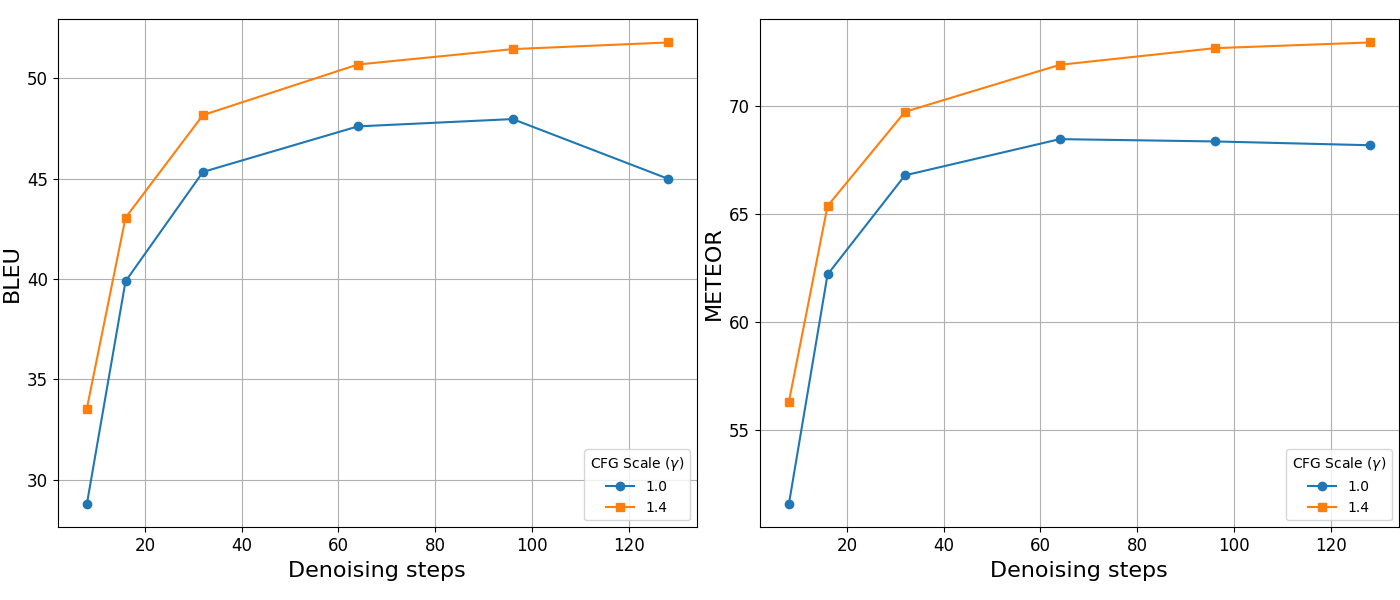}
        \caption{}
        \label{fig:subfig1}
    \end{subfigure}%
    \hfill
    \begin{subfigure}[b]{0.45\textwidth}
        \includegraphics[width=\textwidth]{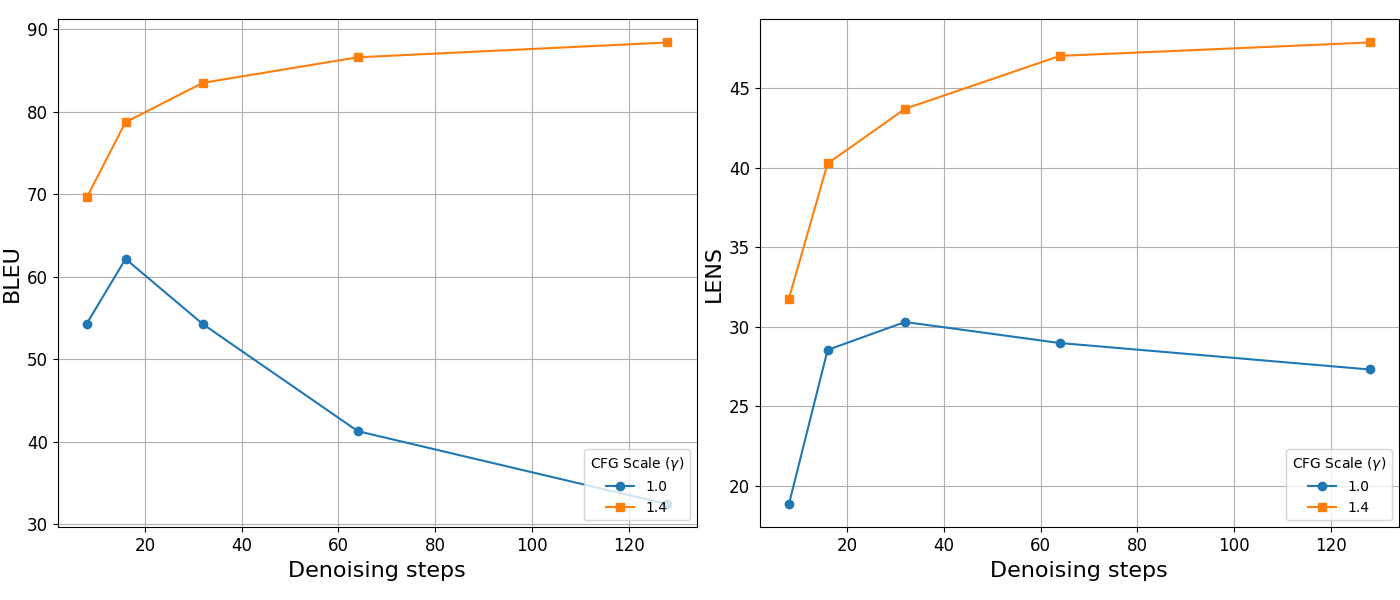}
        \caption{}
        \label{fig:subfig4}
    \end{subfigure}
    \vskip\baselineskip
    \begin{subfigure}[b]{0.45\textwidth}
        \includegraphics[width=\textwidth]{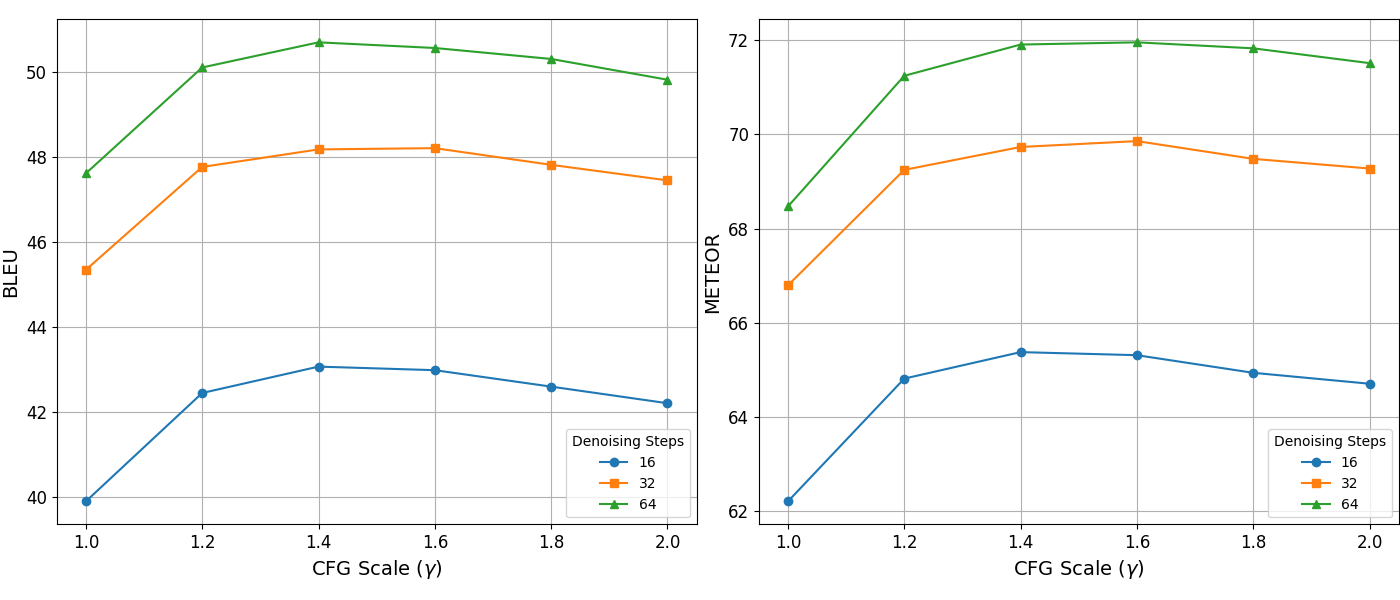}
        \caption{}
        \label{fig:subfig2}
    \end{subfigure}%
    \hfill
    \begin{subfigure}[b]{0.45\textwidth}
        \includegraphics[width=\textwidth]{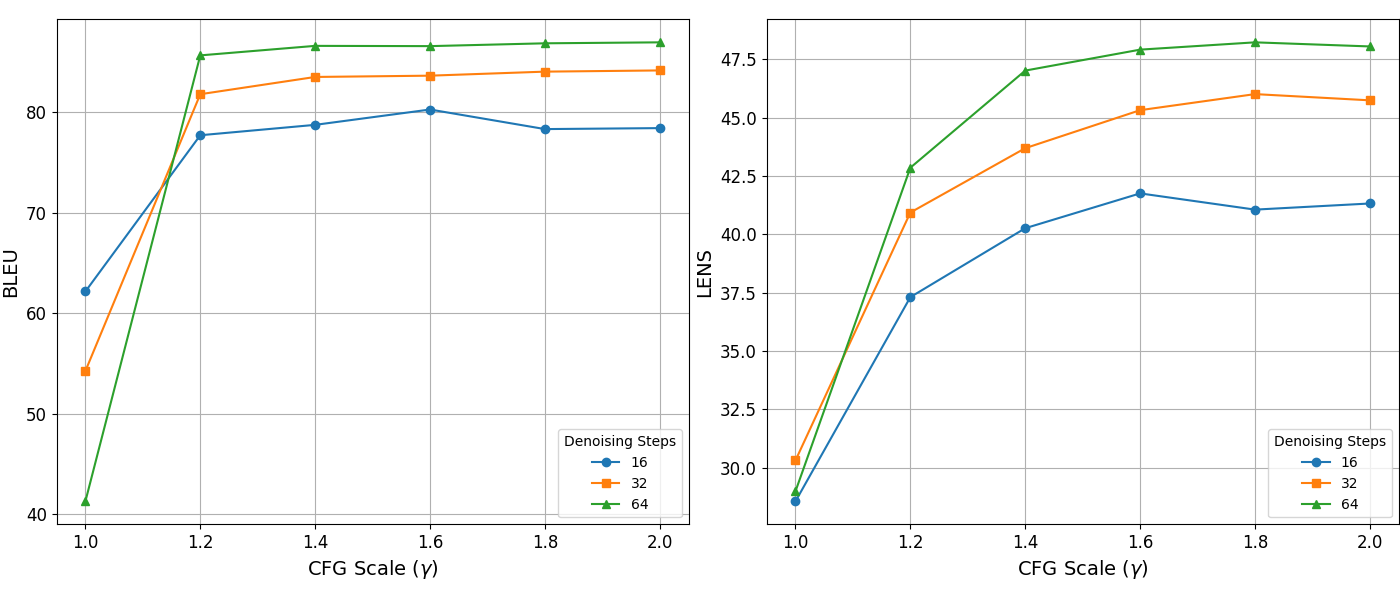}
        \caption{}
        \label{fig:subfig5}
    \end{subfigure}
    \vskip\baselineskip
    \begin{subfigure}[b]{0.45\textwidth}
        \includegraphics[width=\textwidth]{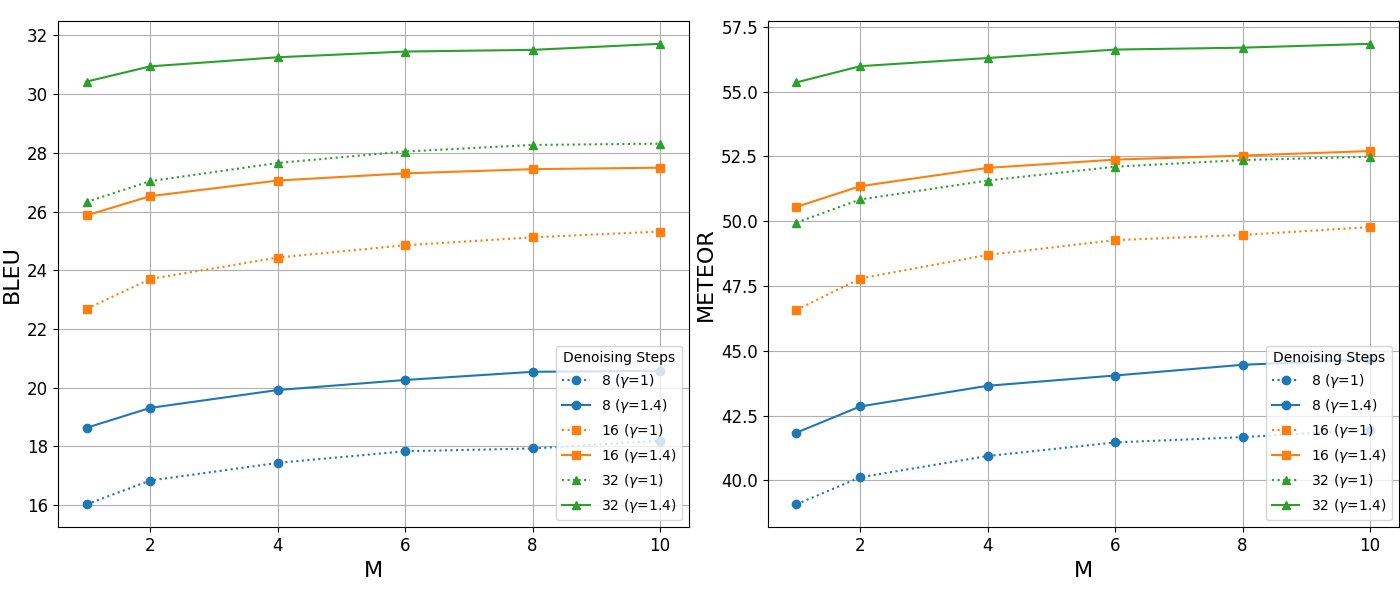}
        \caption{}
        \label{fig:subfig3}
    \end{subfigure}%
    \hfill
    \begin{subfigure}[b]{0.45\textwidth}
        \includegraphics[width=\textwidth]{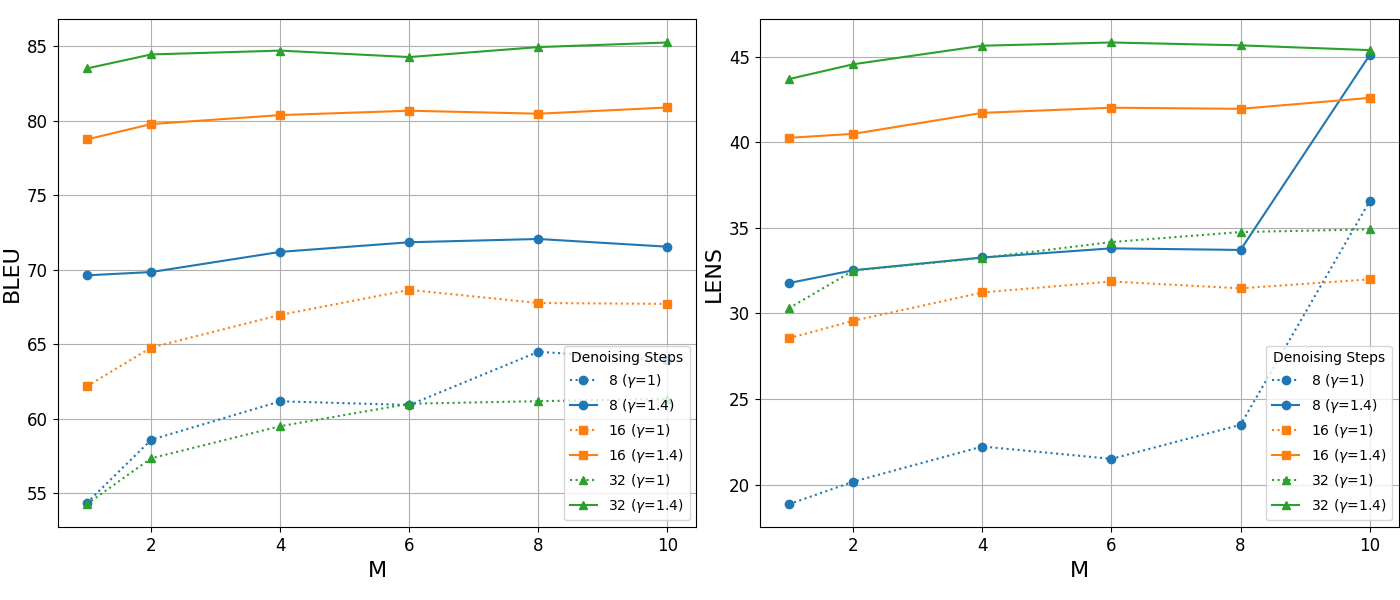}
        \caption{}
        \label{fig:subfig6}
    \end{subfigure}
    \vspace{13pt}
    \caption{Additional plots on inference-time scaling and the effect of CFG scale on the model's performance. Plots (a), (c), and (e) correspond to the PUB-ASV test set, and plots (b), (d), and (f) correspond to the Wikilarge test set.}
    \label{fig:overall}
\end{figure}

\section{Training and Hyperparameter Details} \label{appendix C}
While MDM and GENIE (baseline) are diffusion models, their training and sampling regimes differ significantly. MDMs define the masked diffusion process in the discrete (or token) space while GENIE defines the Gaussian diffusion process in the token embedding (i.e., continuous) space. Also, MDMs are trained in a continuous-time setting (i.e., the diffusion model is trained in the timestep range of $ (0,1] $), which allows us to have flexibility in the discretization of the timestep interval during generation. In contrast, GENIE defines the diffusion process in discrete time, for which it needs to fix the maximum number of discrete timesteps during training. While it is possible to sample in fewer time steps than defined in training (for instance, using the DDIM sampling algorithm \cite{ddim}), GENIE does not employ such methods for text generation, which restricts its sampling process to the same number of time steps as training. 

\subsection{Architecture}
For MDM, we follow \cite{smdm} for the neural network architecture, which is a transformer encoder. Unlike the usual diffusion model architectures, the architecture does not incorporate extra timestep embedding. The model configuration is as follows: 12 layers, 768 hidden dimension size, 12 attention heads, and 3072 intermediate dimension size. The autoregressive baselines follow the transformer decoder architecture. For GENIE, we have an encoder-decoder architecture with the following configuration for both encoder and decoder: 6 layers, 768 hidden and intermediate dimensions, and 128 embedding dimensions.

\begin{table*}[t]
    \centering
    \resizebox{\textwidth}{!}{%
    \begin{tabular}{@{}|c|c|c|c|c|c|c|c|c|c|c|@{}}
    \toprule
    \textbf{Model}                         & \textbf{Dataset} & \textbf{Training steps} & \textbf{Batch size} & \textbf{\begin{tabular}[c]{@{}c@{}}Training \\ Precision\end{tabular}} & \textbf{\begin{tabular}[c]{@{}c@{}}Max Grad \\ Norm\end{tabular}} & \textbf{Warmup steps} & \textbf{Max LR} & \textbf{LR scheduler}                                                                               & \textbf{\begin{tabular}[c]{@{}c@{}}Gradient\\ accumulation\\ steps\end{tabular}} & \textbf{weight decay} \\ \midrule
    \multirow{2}{*}{\textbf{MDM}}          & Bible            & 30000                   & 256                 & \multirow{8}{*}{bf16}                                                  & 1.0                                                               & 1000                  & 3e-4            & \multirow{8}{*}{\begin{tabular}[c]{@{}c@{}}Inverse square root \\ decay with\\ warmup\end{tabular}} & 1                                                                                & 0.01                  \\ \cmidrule(lr){2-4} \cmidrule(lr){6-8} \cmidrule(l){10-11} 
                                           & Wikilarge        & 34000                   & 256                 &                                                                        & 1.0                                                               & 1000                  & 3e-4            &                                                                                                     & 1                                                                                & 0.05                  \\ \cmidrule(r){1-4} \cmidrule(lr){6-8} \cmidrule(l){10-11} 
    \multirow{2}{*}{\textbf{SmolLM2-135M}} & Bible            & 17500                   & 256                 &                                                                        & 1.0                                                               & 500                   & 2e-5            &                                                                                                     & 1                                                                                & 0.01                  \\ \cmidrule(lr){2-4} \cmidrule(lr){6-8} \cmidrule(l){10-11} 
                                           & Wikilarge        & 22000                   & 128                 &                                                                        & 1.0                                                               & 500                   & 4e-5            &                                                                                                     & 2                                                                                & 0.01                  \\ \cmidrule(r){1-4} \cmidrule(lr){6-8} \cmidrule(l){10-11} 
    \multirow{2}{*}{\textbf{SmolLM2-360M}} & Bible            & 18000                   & 128                 &                                                                        & 1.0                                                               & 1000                  & 5e-6            &                                                                                                     & 2                                                                                & 0.05                  \\ \cmidrule(lr){2-4} \cmidrule(lr){6-8} \cmidrule(l){10-11} 
                                           & Wikilarge        & 14000                   & 128                 &                                                                        & 1.0                                                               & 500                   & 3e-5            &                                                                                                     & 2                                                                                & 0.1                   \\ \cmidrule(r){1-4} \cmidrule(lr){6-8} \cmidrule(l){10-11} 
    \multirow{2}{*}{\textbf{Qwen2.5-0.5B}} & Bible            & 16000                   & 128                 &                                                                        & 1.0                                                               & 500                   & 1e-6            &                                                                                                     & 2                                                                                & 0.1                   \\ \cmidrule(lr){2-4} \cmidrule(lr){6-8} \cmidrule(l){10-11} 
                                           & Wikilarge        & 32000                   & 128                 &                                                                        & 1.0                                                               & 500                   & 2e-6            &                                                                                                     & 2                                                                                & 0.1                   \\ \midrule
    \multirow{2}{*}{\textbf{GENIE}}        & Bible            & 60000                   & 512                 & \multirow{2}{*}{fp32}                                                  & -                                                                 & 7200                  & 1e-4            & \multirow{2}{*}{\begin{tabular}[c]{@{}c@{}}Linear schedule\\  with warmup\end{tabular}}             & 1                                                                                & \multirow{2}{*}{0.0}    \\ \cmidrule(lr){2-4} \cmidrule(lr){6-8} \cmidrule(lr){10-10}
                                           & Wikilarge        & 50000                   & 256                 &                                                                        & -                                                                 & 7200                  & 5e-5            &                                                                                                     & 1                                                                                &                       \\ \bottomrule
    \end{tabular}%
    }
    \vspace{4pt}
    \caption{Hyperparameter settings for the experiments reported in the paper.}
    \label{table: hyp}
    \vspace{6pt}
\end{table*}

\begin{table*}[t]
    \centering
    \setlength{\tabcolsep}{30pt}
    \resizebox{0.8\textwidth}{!}{%
    \begin{tabular}{@{}ccccc@{}}
    \toprule
    \multicolumn{5}{c}{\textbf{PUB-BBE}}                                                                               \\ \midrule
    \textbf{Steps} & \textbf{BLEU}          & \textbf{ROUGE-L}       & \textbf{METEOR}        & \textbf{BERTScore}     \\
    8              & $ 19.796_{\pm 0.048} $ & $ 46.741_{\pm 0.064} $ & $ 43.522_{\pm 0.074} $ & $ 88.663_{\pm 0.025} $ \\
    16             & $ 27.045_{\pm 0.039} $ & $ 53.256_{\pm 0.033} $ & $ 51.810_{\pm 0.046} $ & $ 90.778_{\pm 0.028} $ \\
    64             & $ 33.252_{\pm 0.032} $ & $ 46.901_{\pm 0.056} $ & $ 43.648_{\pm 0.077} $ & $ 88.751_{\pm 0.024} $ \\ \midrule
    \multicolumn{5}{c}{\textbf{PUB-ASV}}                                                                               \\ \midrule
    8              & $ 35.075_{\pm 0.035} $ & $ 60.969_{\pm 0.046} $ & $ 58.010_{\pm 0.041} $ & $ 91.196_{\pm 0.020} $ \\
    16             & $ 44.071_{\pm 0.049} $ & $ 67.668_{\pm 0.037} $ & $ 66.420_{\pm 0.050} $ & $ 93.102_{\pm 0.019} $ \\
    64             & $ 51.103_{\pm 0.032} $ & $ 72.661_{\pm 0.039} $ & $ 72.405_{\pm 0.035} $ & $ 94.587_{\pm 0.036} $ \\ \bottomrule
    \end{tabular}
    }
    \vspace{4pt}
    \caption{Additional results on the Bible dataset using static word vector embeddings. All results are computed with a guidance scale of $1.4$.}
    \label{table: add_results}
    \vspace{6pt}
\end{table*}

\section{Additional Experiments with Static Word Vector Embeddings} \label{appendix D}
Alongside the usage of semantic contextual embeddings from sentence embedding models, we also performed experiments with static word vector embeddings. To perform this experiment, we take fasttext embeddings of each word in the sentence and average them to use as a sentence embedding. Table \ref{table: add_results} shows the results on the bible dataset. We observe that the overall performance is not very far behind that of the contextual sentence embeddings. 

Contextual sentence embeddings are better at capturing the semantic content of a sentence than averaged static word vector embeddings. However, they also tend to embed style along with the semantics of the sentence. In contrast, static word vector embeddings are style agnostic but do not capture semantics very well. As both settings have their own flaws, it can be an interesting exploration of how both can be combined. Another way can be to fine-tune contextual sentence embeddings to be style agnostic, given parallel style transfer data, which may potentially improve the performance.

\section{Additional Plots on Inference-time Scaling} \label{appendix E}
In Figure \ref{fig:overall}, we include the plots on inference-time scaling and the effect of CFG scale for PUB-ASV and the Wikilarge test sets (similar to PUB-BBE plots in Figure 3 of the main content). We mostly observe trends similar to those of the PUB-BBE dataset. Without CFG, the performance on the WikiLarge dataset drops much more quickly than the Bible test sets as we scale denoising steps. Interestingly, we observe a sharp spike in simplicity scores with 8 denoising steps and 10 candidate generations with the SVDD algorithm (as seen in Figure \ref{fig:subfig6}). 

\begin{figure*}[h]
\centering
    \begin{examplebox}[\small]
    \textbf{PUB-BBE Generated outputs:} \\
    \textbf{Input:} And Jehovah was with Judah; and drove out the inhabitants of the hill-country; for he could not drive out the inhabitants of the valley, because they had chariots of iron. \\ \\
    \textbf{Reference:} And the Lord was with Judah; and he took the hill-country for his heritage; but he was unable to make the people of the valley go out, for they had war-carriages of iron. 

    \begin{itemize}
        \setlength\itemsep{0.5em}
        \item \textbf{MDM} ($M=4$, $\gamma=1.4$, \textit{steps} = 64): And the Lord was with Judah, driving out the people of the hill-country; for he was not able to overcome the people of the valley, because they had war-carriages of iron.
        \item \textbf{MDM} ($M=1$, $\gamma=1.0$, \textit{steps} = 64): And the Lord was with Judah, driving out the people of the hill-country; for he was not able to send the people of the valley, because they had their carriages of iron
        \item \textbf{GENIE:} and the lord was with judah, driving out the people of the hill-country ; for they had a desire not able to send out the people of the valley, because they were war - of iron.
        \item \textbf{SmolLM2-135M:} And the Lord was with Judah, and made them the heritage of the nations, \textcolor{red}{because of the fighting-men of the hill-country;} for they were able to overcome the fighting men of the valley, because they had the use of iron in their armies.
        \item \textbf{SmolLM2-360M:}And the Lord kept back from Judah all those who had taken cover in the hills: for he could not get the hill-country to go away from him, because they had iron chariots.
        \item \textbf{Qwen2.5:} And the Lord was with Judah; and the men of the hill-country were driven out; because he was not able to drive out the inhabitants of the valley, because of the \textcolor{red}{iron-horned men}.
    \end{itemize}
    
    \vspace{3pt}
    \hrule
    \vspace{4pt}
    
    \textbf{PUB-ASV Generated outputs:} \\
    \textbf{Input:} The raiders went out from the camp of the Philistines in three divisions. One division turned on the road to Ophrah toward the land of Shual. \\ \\
    \textbf{Reference:} And the spoilers came out of the camp of the Philistines in three companies: one company turned unto the way that leadeth to Ophrah, unto the land of Shual.

    \begin{itemize}
        \setlength\itemsep{0.5em}
        \item \textbf{MDM} ($M=4$, $\gamma=1.4$, \textit{steps} = 64): And the destroyers went out from the camp of the Philistines in three hosts: and one course turned by the way of Ophrah, unto the land of Shual.
        \item \textbf{MDM} ($M=1$, $\gamma=1.0$, \textit{steps} = 64): And the archers went out from the camp of the Philistines in three companies: and one course turned toward the way of Ophrah, unto the land of Shual.
        \item \textbf{GENIE:} and the raiders went forth out of the camp of the philistines in three divisions ; one division turned the way to ophrah toward the land of shual.
        \item \textbf{SmolLM2-135M:} And the robbers fled from the camp of the Philistines unto three companies, which turned to the way to Ophrah, toward the land of Susan.
        \item \textbf{SmolLM2-360M:}And the camp of the Philistines had passed forth as three companies; and one passed the way to Ophrah, \textcolor{red}{and another went up into the wilderness by the way to Shalem.}
        \item \textbf{Qwen2.5:} And the raiders went forth from the camp of the Philistines in three divisions. One division turned on the way to Ophrah, to the land of Shaual, \textcolor{red}{and overthrew their cities.}
    \end{itemize}

    \vspace{3pt}
    \hrule
    \vspace{4pt}
    
    \textbf{Wikilarge Generated outputs:} \\
    \textbf{Input:} if there are no strong land use controls , buildings are built along a bypass , converting it into an ordinary town road , and the bypass may eventually become as congested as the local streets it was intended to avoid. \\ \\
    \textbf{Reference:} if there are no strong land use controls , the bypass may eventually become as congested as the local streets .

    \begin{itemize}
        \setlength\itemsep{0.5em}
        \item \textbf{MDM} ($M=4$, $\gamma=1.4$, \textit{steps} = 64): if there are no strong land use controls , buildings are built along a bypass , changing it into an ordinary town road .
        \item \textbf{MDM} ($M=1$, $\gamma=1.0$, \textit{steps} = 64): the bypass .
        \item \textbf{GENIE:} if there are no strong land use controls, buildings are built along a bypass and turn it into an ordinary town road. the bypass may also go, and the it may eventually become as congested as the local streets it intended to avoid.
        \item \textbf{SmolLM2-135M:} In a city without strong land use controls , buildings are built along a bypass , changing it into an ordinary town road , and the bypass may eventually become as congested as the local streets it was intended to avoid .
        \item \textbf{SmolLM2-360M:}If there are no strong land use controls , buildings are built along a bypass , turning a road into a street .
        \item \textbf{Qwen2.5:} In addition , if there are no strong land use controls , buildings are built along a bypass , converting it into an ordinary town road .

    \end{itemize}
    
    \end{examplebox}
\caption{Generated output examples for all the test sets considered in the evaluation. Out-of-context (and sometimes hallucinatory) behaviour in autoregressive LM generations is marked with \textcolor{red}{red}. For Wikilarge generations, we see vanilla MDMs without inference-time scaling outputs an extremely small and inadequate output. As discussed in the main content, this may happen due to shorter sentences being mistaken for simplicity.}
\vspace{0.2cm}
\label{fig: generated-examples-supp}
\end{figure*}

\end{document}